\documentclass{article}

     \PassOptionsToPackage{numbers, compress}{natbib}
     \usepackage[final]{neurips_2025}

\usepackage[utf8]{inputenc} % allow utf-8 input
\usepackage[T1]{fontenc}    % use 8-bit T1 fonts
\usepackage{url}            % simple URL typesetting
\usepackage{booktabs}       % professional-quality tables
\usepackage{amsfonts}       % blackboard math symbols
\usepackage{amssymb}
\usepackage{amsmath}
\usepackage{nicefrac, xfrac}       % compact symbols for 1/2, etc.
\usepackage{microtype}      % microtypography
\usepackage{newunicodechar}
\newunicodechar{⚔}{\textdaggerdbl} % fallback for sword symbol
\usepackage{xcolor}         % colors
\usepackage[ruled,linesnumbered,boxed,commentsnumbered]{algorithm2e}
\usepackage{algpseudocode}  % 常用算法宏包
\usepackage{multirow}
\usepackage{makecell}
\usepackage{graphicx}
\usepackage{tabulary}
\usepackage{longtable} 
\usepackage{caption}
\usepackage{wrapfig}
\usepackage{arydshln}
\usepackage{enumitem}
\DontPrintSemicolon
\usepackage{booktabs}
\definecolor{citecolor}{HTML}{0071bc}
\definecolor{shadecolor}{rgb}{0.9,0.9,0.9}
\definecolor{pink}{HTML}{e78e8c}

\usepackage[colorlinks, linkcolor=red, colorlinks, anchorcolor=blue, citecolor=citecolor]{hyperref}
\usepackage{url}
\usepackage{graphicx}
\usepackage{booktabs}
\usepackage{enumitem}

\usepackage{amssymb}
\usepackage{amsmath}
\usepackage{wrapfig}
\usepackage{lipsum}
\usepackage{graphicx}
\usepackage{graphics}
\usepackage{adjustbox}
\usepackage{setspace}
\usepackage{multirow}
\usepackage{multicol}
\usepackage{colortbl}
\usepackage{circledsteps}
\usepackage{tabularx}
\usepackage{subcaption} % For subfigures
\usepackage{soul}% for underlines
\usepackage[symbol]{footmisc}
\usepackage{wrapfig}
\usepackage{listings}
\usepackage{tcolorbox}
\usepackage{xcolor}
\usepackage{enumitem}

\definecolor{myblue}{HTML}{4A90E2}
\definecolor{mygreen}{HTML}{50C878}
\definecolor{myred}{HTML}{D9534F}
\definecolor{mygray}{rgb}{0.5,0.5,0.5}
\definecolor{myorange}{HTML}{FF7A33}
\definecolor{lightgray}{rgb}{0.9, 0.9, 0.9}

\usepackage{etoolbox}

\usepackage{graphicx}
\usepackage{bm}
\usepackage{wrapfig}
\usepackage{threeparttable}
\usepackage{epsfig}
\usepackage{caption}
\usepackage{color}
\usepackage{colortbl}
\usepackage{comment}
\usepackage{algpseudocode} % Added by Luis
\usepackage{breqn} % Added by Luis
\usepackage{dsfont}
\usepackage{float}
\usepackage{booktabs}

\newcommand{\secvsabove}{\vspace{-1mm}}
\newcommand{\secvsbelow}{\vspace{-1.5mm}}

\newcommand{\paravs}{\vspace{-2mm}}

\newcommand{\ourmethod}[1]{\textsc{Sparta Alignment}}

\title{\ourmethod{}: Collectively Aligning Multiple Language Models through Combat}
\author{Yuru Jiang\thanks{equal contribution} \ \textsuperscript{1} \ \ \ Wenxuan Ding\footnotemark[1] \ \textsuperscript{2} \ \ \ Shangbin Feng\footnotemark[1] \ \textsuperscript{3} \ \ \ Greg Durrett\textsuperscript{2} \ \ \ Yulia Tsvetkov\textsuperscript{3} \\
\textsuperscript{1}Zhejiang University \ \ \ \ \ \textsuperscript{2}New York University \ \ \ \ \ \textsuperscript{3}University of Washington \\
\texttt{3220102689@zju.edu.cn} \ \ \ \texttt{wd2403@nyu.edu} \ \ \ \texttt{shangbin@cs.washington.edu}
}

\begin{document}

\maketitle

\begin{abstract}
  We propose \ourmethod{}, an algorithm to collectively align multiple LLMs through competition and combat. To complement a single model's lack of diversity in generation and biases in evaluation, multiple LLMs form a ``sparta tribe'' to compete against each other in fulfilling instructions while serving as judges for the competition of others. For each iteration, one instruction and two models are selected for a duel, the other models evaluate the two responses, and their evaluation scores are aggregated through a adapted elo-ranking based reputation system, where winners/losers of combat gain/lose weight in evaluating others. The peer-evaluated combat results then become preference pairs where the winning response is preferred over the losing one, and all models learn from these preferences at the end of each iteration. \ourmethod{} enables the self-evolution of multiple LLMs in an iterative and collective competition process. Extensive experiments demonstrate that \ourmethod{} outperforms initial models and 4 self-alignment baselines across 10 out of 12 tasks and datasets with 7.0\% average improvement. Further analysis reveals that \ourmethod{} generalizes more effectively to unseen tasks and leverages the expertise diversity of participating models to produce more logical, direct and informative outputs.\footnote{Resources available at \href{https://github.com/yurujiang2003/sparta}{https://github.com/yurujiang2003/sparta}.}
\end{abstract}

\secvsabove
\section{Introduction}
\secvsbelow
Aligning large language models (LLMs) has become a critical step of LLM post-training, steering LLMs for specialized skills~\citep{fengknowledge,feng2024model}, helpful responses~\citep{bai2022training}, and alignment with human values~\citep{hendrycks2023aligningaisharedhuman,christiano2017,sun-etal-2024-aligning}. Existing alignment recipes would require external signals to learn from (e.g., reward models~\citep{ibarz2018reward,lindner2021information,chen2025pal} or preference pairs~\citep{song2024preference,ibarz2018reward,dpo,meng2024simpo}) while high-quality supervision is often scarce and costly~\citep{bai2022training,ouyang2022training}. As such, recent research investigates \emph{self-alignment}, where the LLM itself serves as its own judge~\citep{ko2025sera}, reward model~\citep{wu2023fine,cui2025processreinforcementimplicitrewards}, or supervision signal~\citep{kiruluta2025selfsupervisedreinforcementlearningapproach,song2024preference}, without any external information. These approaches have demonstrated gains on instruction following and reasoning problems, initiating a paradigm of self-improving language models~\citep{zelikman2022star,yuan2025selfrewardinglanguagemodels,madaan2024self,wu2024selfplaypreferenceoptimizationlanguage,chen2024self}.

However, we posit that the recipe of ``one model aligning itself''  might not be reliable for two reasons:
\begin{itemize}[leftmargin=*]
    \item  \emph{A single model struggles to reliably judge its own generation}. There is ample evidence on LLM self-bias~\citep{xu2024prideprejudicellmamplifies,feuer2025styleoutweighssubstancefailure,wang2025ragenunderstandingselfevolutionllm}, where one model consistently favors its own response and reinforces its own priors/biases in LLM-as-a-judge evaluations~\citep{ye2025learning}. Impact of the biased self-judgment will be even more pronounced for tasks and instructions with underpinning human values and cultural implications~\citep{shi2024culturebank,naous2023having,rao2024normad}, propagating and reinforcing the flaws.
    \item \emph{A single model struggles to reliably generate diverse responses to learn from}. RL and preference learning recipes benefit from a wide spectrum of responses with varying quality and reward scores~\citep{yao2024varyingshadeswrongaligning}. A single model, despite sampling, mostly generates homogeneous responses with similar styles and failure patterns~\citep{slocum2025diverse,zelikman2022star}, struggling to provide the diversity and distinguishability of responses for robust preference learning.
\end{itemize}
Therefore, the single model itself becomes the bottleneck for self-evolution beyond its training priors, existing strengths and weaknesses, as well as its inherent biases.

To this end, we propose \ourmethod{}, an algorithm to collectively align multiple language models. To complement each model's limitation in \emph{judging} and \emph{generating}, multiple LLMs form a ``sparta tribe'' to compete against each other~\cite{li2024fightladderbenchmarkcompetitivemultiagent,xiong2023examining,du2024improving} at fulfilling instructions and at the same time serve as judges for other LLMs. Specifically, \ourmethod{} takes a pool of LMs $\mathcal{M}^0$ and a set of instructions $\mathcal{X}$ as input. For each iteration $t$, we repeatedly sample an instruction $x \in \mathcal{X}$ and design a \emph{match-making system} to select two models $M_i^t$ and $M_{i'}^t$ for ``combat'', i.e., dueling to better respond to $x$. All the other models $\mathcal{M}^t \setminus \{M_i^t, M_{i'}^t\}$ then judge their responses and the aggregated judgment determines who won the combat, resulting in a preference pair such as $(x, M_i^t(x) \prec M_{i'}^t(x))$. Motivated by game theory~\citep{ebtekar2021elo}, the judgment aggregation is weighted by the \emph{reputation scores} of models, where the winning models increase and the losing models decrease after combat. At the end of each iteration, all of $\mathcal{M}^t$ go through preference learning~\citep{schulman2017proximal,guo2024direct} with the collection of preference pairs in this iteration. After iterations of \ourmethod{}, a few models in $\mathcal{M}^t$ would emerge as the strongest with the highest reputation scores for inference and deployment. \ourmethod{} uniquely enables the self-evolution~\citep{zuo2025ttrltesttimereinforcementlearning} of LLMs in a collective competition process, alleviating the limitations of a single model.

Extensive experiments on 12 datasets show that \ourmethod{} consistently improves LLMs across 10 of the 12 tasks, outperforming the strongest baseline by 7\% on average. \ourmethod{} achieves the highest generalization accuracy in 2 out of 3 categories on challenging mathematical benchmarks, demonstrating robust generalization~\cite{chih2024magicore} beyond~$\mathcal{X}$. \ourmethod{} resembles the social stratification theory~\citep{magee20088,magee2008self,sauder2012status}, where models in the Sparta tribe show different ``classes'' of reputation scores at the end of alignment. Further analysis reveals that the ``sparta tribe'' benefits from having more members and that \ourmethod{} benefits from a diverse set of models with different expertise as $\mathcal{M}$ for improved generation diversity and more distinguishable preference pairs. 

\begin{figure}[t]
    \centering
    \includegraphics[width=1\linewidth]{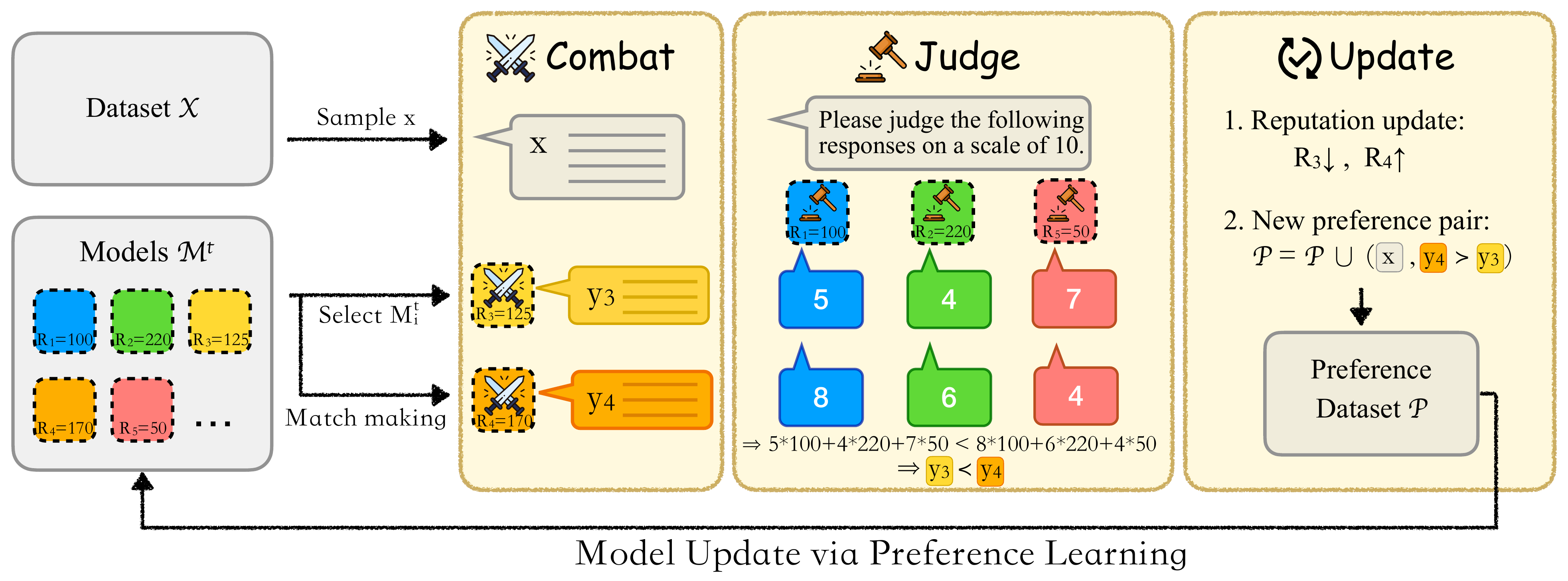}
    % \vspace{-0.1in}
    \caption{Overview of \ourmethod{}, an algorithm collectively aligning multiple models via combat. \ourmethod{} requires a dataset $\mathcal{X}$ and a pool of models $\mathcal{M}^t$. For each iteration $t$, we repeatedly sample a prompt $x$ from $\mathcal{X}$. For $x$, we first select a model $M_i^t$ and then select its opponent $M_{i'}^t$ with our Match-Making Strategy ($i = 3, i' = 4$ in the example). We then employ three steps: (1) Combat: the model pair generate responses $y_i, y_{i'}$ to $x$. (2) Judge: other models in the pool act as judges, generating their scores to $y_i, y_{i'}$. (3) Update: update the reputation scores of $M_i^t$ and $M_{i'}^t$ based on scores from Judge phase and create a new preference for the preference dataset $\mathcal{P}$. At the end of each iteration, we align models via preference learning on $\mathcal{P}$.}
    \label{fig:overview}
\end{figure}
\secvsabove
\section{Methodology}\label{method}
\secvsbelow
\paragraph{Overview} \textsc{Sparta Alignment} assumes access to a pool of LLMs, denoted as $\mathcal{M}^0 = \{M_k^0\}_{k=1}^{m}$ where $0$ denotes the 0-th iteration, and a dataset $\mathcal{X}$. The models evolve iteratively for $\mathcal{T}$ iterations. In iteration $t$, we repeatedly sample $x \in \mathcal{X}$, select two models $M_i^t, M_{i'}^t$ for head-to-head \textit{combat} via our Match-Making System, and generate responses $M_i^t(x), M_{i'}^t(x)$. The other models $\mathcal{M}^t \setminus \{M_i^t, M_{i'}^t\}$ then act as judges and evaluate $M_i^t(x)$ and $M_{i'}^t(x)$ to produce scores. These scores are aggregated using a weighted scheme based on each judge’s reputation $R_k$, yielding a collective preference signal. The reputation system then updates the reputation $R_i$ and $R_{i'}$ by rewarding the winner and penalizing the loser, and we add the preference pair to the current iteration's preference dataset $\mathcal{P}$. At the end of each iteration, the accumulated preference pairs in $\mathcal{P}$ are used to fine-tune all models via Direct Preference Optimization (DPO) \citep{rafailov2024direct}, progressively aligning multiple language models over the iterations.
We present an overview of \ourmethod{} in Figure \ref{fig:overview} and Algorithm \ref{alg:main}, followed by detailed explanation of the three key components in the algorithm: \emph{Match-Making System}, \emph{Judgment Aggregation}, and \emph{Reputation System}.
\SetKwInOut{KwIn}{Input}
\SetKw{Kwhp}{Hyperparameters:}
\begin{algorithm}[t]
    \SetAlgoLined
    \small
    \caption{Sparta Alignment}\label{alg:main}
\textbf{Input:}{\textup{
    Initial models $\mathcal{M}^0 = \{M_i^0\}_{i=1}^{m}$, with attributes $R_i$, where $R_i$ is model's reputation dynamically updated by combat results, 
    Dataset $\mathcal{X}$;
    Max iteration $\mathcal{T}$.
}}
    
    \For{$t = 1$ \KwTo $\mathcal{T}$}{
        Preference dataset $\mathcal{P} \gets \emptyset$\;
        \For{$x \in \mathcal{X}$}{
            Select a model $M_i^t$ from $\mathcal{M}^t$\; 
            Select its opponent $M_{i'}^t = \text{Match-Making}(M_i^t)$\;
            
            Combat: both models generate a response to $x$, $M_i^t(x)$ and $M_{i'}^t(x)$ \;
            
            \For{$M_k^t \in \mathcal{M}^t \setminus \{M_i^t, M_{i'}^t\}$}{
                Evaluate both responses: $s_i^{(k)} \gets \mathcal{J}(x, M_i^t(x), M_k^t)$, $s_{i'}^{(k)} \gets \mathcal{J}(x, M_{i'}^t(x), M_k^t)$, where $\mathcal{J}(x, M_i^t(x), M_k^t)$ denotes $M_k^t$ evaluating response $M_i^t(x)$  \;
            }

            $\bar{s_i} = \text{Aggregate}(\{s_i^{(k)}\}, \{R_k\}\})$, $\bar{s_{i'}} = \text{Aggregate}(\{s_{i'}^{(k)}\}, \{R_k\}), M_k^t \in \mathcal{M}^t \setminus \{M_i^t, M_{i'}^t\}$\;
            
            \textbf{if} $\bar{s_i} > \bar{s_{i'}}$ \textbf{then} new preference pair $\mathcal{P} \gets \mathcal{P} \cup \{x, M_i^t(x) \succ M_{i'}^t(x)\}$\\
            \textbf{else} new preference pair $\mathcal{P} \gets \mathcal{P} \cup \{x,  M_{i'}^t(x) \succ M_i^t(x)\}$\\
            %\textbf{else} pass\;
            
            $R_i, R_{i'} \gets \text{Reputation-Update}(R_i, R_{i'}, \bar{s_i}, \bar{s_{i'}})$ \;
        }
        
        \For{$M_i^t \in \mathcal{M}^t$}{
            $M_i^{t+1} \gets \mathrm{DPO}(M_i^t, \mathcal{P})$
        }
    }
    \textbf{Output:} Better Models $\mathcal{M}^{\mathcal{T}} = \{M_i^{\mathcal{T}}\}_{i=1}^{m}$
\end{algorithm}

\paravs
\paragraph{Match-Making System}\label{match-making}
Competitions between models of similar strength levels would help distinguishability and produce more meaningful preference signals \citep{li2024autobencher}. To this end, \ourmethod{} uses a probabilistic match-making strategy to select two LLMs to \emph{combat} for better fulfilling an instruction.  For a given data instance $x$, the match-making process proceeds as follows:
We begin by selecting the first model randomly, $M_i^t$, in the model pair from the model pool $\mathcal{M}^t$. The opponent model $M_{i'}^t$ is chosen partly based on reputation scores: with predefined probability $\alpha$, $M_{i'}^t$ is randomly selected from the pool. Otherwise, with probability $1-\alpha$, 
$M_{i'}^t$ is selected from the top-$k$ models whose reputation scores are closest to that of $M_i^t$. We denote this process as $M_{i'}^t = \text{Match-Making($M_i^t$)}$, ensuring a balance between exploring diverse models and exploiting those with similar reputation scores.
\paravs
\paragraph{Judgment Aggregation}\label{aggregate}
To obtain reliable and comprehensive preference signals, \ourmethod{} aggregates evaluation scores from all remaining models in the pool. For each data instance \(x\), the selected models \(M_i^t\) and \(M_{i'}^t\) generate responses \(M_i^t(x)\) and \(M_{i'}^t(x)\), respectively. These responses are then evaluated by all other models \(M_k^t \in \mathcal{M}^t \setminus \{M_i^t, M_{i'}^t\}\), each acting as an independent judge and assign score  $s_i^{(k)}=\mathcal{J}(x, M_i^t(x), M_k^t)$ to \(M_i(x)\) on a scale of ten, where $\mathcal{J}$ denotes LLM-as-a-judge. Then, the overall score $\bar{s_i}$ is calculated using reputation-weighted aggregation:
\begin{align*}
\bar{s_i} & = \mathrm{Aggregate}(\{s_i^{(k)}\}, \{R_k\}) = \frac{\sum_{k \in \mathcal{M}^t \setminus \{M_i^t, M_{i'}^t\}}  R_k s_i^{(k)}}{\sum_{k \in \mathcal{M}^t \setminus \{M_i^t, M_{i'}^t\}}  R_k}
% , \text{where }M_k \notin \{M_i, M_{i'}\} 
\end{align*}
and the score $\bar{s_{i'}}$ for $M_{i'}^t(x)$ is calculated similarly. Specifically, $R_k$ is the reputation score of model $M_k^t$ in the pool, reflecting its perceived credibility among peers. 
Through reputation weighting, we aim for a more reliable judgment by aggregating the collective insights of all models. 

\paravs
\paragraph{Reputation System}\label{reputation_system}  
\ourmethod{} captures each model's strength from the perspective of its peers, enabling more accurate and stable reputation updates while reducing inconsistency in preference labels~\citep{wang2025creamconsistencyregularizedselfrewarding,zhou2025selfconsistencyinternalrewardmodels}.

At initialization, all models are assigned the same reputation score to ensure a fair starting point. During training, these scores are iteratively updated based on two key factors: the aggregated judgment by other models and each model’s own reputation deviation, which measures the fluctuation of reputation over a sliding window. This mechanism enables the system to progressively distinguish models by their demonstrated performance while accounting for their recent reputation changes.
Formally, the reputation $R_{i}$ is updated as follows:
\begin{equation} 
    R_{i} \gets R_{i} + \textcolor{purple}{\kappa \cdot (\bar{s}_i - \bar{s}_{i'})} \cdot \textcolor{blue}{\tanh(\sigma_{i})} \cdot \textcolor{teal}{\max\left( \left| \Phi(z_i) - \Phi(z_{i'}) \right|, \epsilon \right)}.
    \label{eq:reputation}
\end{equation}
We denote the process as $R_i, R_{i'} \gets \text{Reputation-Update}(R_i, R_{i'}, \bar{s_i},  \bar{s_{i'}})$ and explain the three underlying principles below:
\begin{itemize}[leftmargin=*]    
    \item \emph{\textcolor{purple}{Larger Score Gap Amplifies Impact}:}  
    The score gap \((\bar{s}_i - \bar{s}_{i'})\) reflects the collective judgment by peer models. A larger aggregated score gap, which implies greater distinction in model capability, leads to proportionally larger reputation update, where $\kappa$ is the scaling factor.
    
    \item \emph{\textcolor{blue}{Deviation-Guided Updates}:}  The reputation score update is moderated based on the variance of the model reputation in recent iterations, which we measure by deviation \(\sigma_i\).
    Specifically, we consider the $N$ most recent iterations, where the reputation change in each iteration is $\delta_t = R_{i}^{(t+1)} - R_{i}^{(t)}$, for $t=1, \dots, N$.
    Then the deviation is calculated as:
    \begin{align*} 
        \sigma_{i} \gets \max\left( \sqrt{\frac{1}{N-1} \sum_{t=1}^{N} (\delta_t - \bar{\delta})^2}, \; \sigma_{\text{min}} \right),
        \quad \text{where} \ \bar{\delta} = \frac{1}{N} \sum_{t=1}^{N} \delta_t
        \label{eq:deviation}
    \end{align*}
    This mechanism ensures higher deviations for unstable model reputations, allowing faster adjustments, while stable models are updated more conservatively. The lower bound \(\sigma_{\text{min}}\) prevents degeneration of the reputation updates.
    
    The scaling term \(\tanh(\sigma_{i})\) in Equation~\ref{eq:reputation} adjusts the update magnitude based on the its stability of reputation scores. For stable models (small $\sigma_i$), updates are small to avoid overreaction, while for unstable ones (large $\sigma_i$), updates are larger for faster convergence. The \(\tanh(\cdot)\) function naturally caps the scaling within \((0, 1)\), preventing excessive updates while providing a smooth and adaptive transition between stable and unstable scenarios.
    \item \emph{\textcolor{teal}{Greater Gains for Defeating Stronger Opponents}:} Motivated by game theory~\citep{ebtekar2021elo}, we employ the term \(\max\left( \left| \Phi(z_i) - \Phi(z_{i'}) \right|, \epsilon \right)\) to capture the relative strength between competing models. Here, \(z_i = \dfrac{R_i - R_{i'}}{\sqrt{\sigma_i^2 + \sigma_{i'}^2}}\) measures the reputation difference normalized by their deviations, and \(\Phi(\cdot)\) denotes the standard normal cumulative distribution function (CDF),  which is used to estimate the model's most probable win rate based on its current reputation and deviation.
    This factor grants larger reputation gain for defeating stronger opponents, rewarding unexpected victories, and vice versa.
    The lower bound $\epsilon$ prevents the update from vanishing when the reputation difference is small, thereby avoiding stagnation and ensuring active reputation updates even in near-tie scenarios.
\end{itemize}
We provide further details about an optional reputation reweighting mechanism in Appendix~\ref{reweight}. 

\secvsabove
\section{Experiment Settings}\label{experiment}
\secvsbelow
\paragraph{Models and Implementation} 
In the main experiments, we initialize the model pool $\mathcal{M}^0$ as 10 \textbf{Qwen2.5-7B-Instruct} models~\citep{qwen2025qwen25technicalreport} independently fine-tuned on different data domains in Tulu-v2~\citep{ivison2023camels} (details in Appendix~\ref{sft-data}), resulting in a pool of models with diverse skills and expertise for collective alignment.

For \ourmethod{}, we set the number of prompts per iteration to $1000$, number of iterations $\mathcal{T}=8$, $\alpha = 0.6$, a top-k threshold of $k = 5$, and $\kappa = 1$. 
At the end of each iteration, all models are fine-tuned via DPO~\citep{rafailov2024direct} for $1$ epoch with a starting learning rate of $1e-6$ and an effective batch size of 1, with the same LoRA configuration in SFT phase in Appendix~\ref{sft-data}. Experiments are performed on a cluster with 16 A100 GPUs with 40 GB memory.

\paravs
\paragraph{Baselines}
 We compare \ourmethod{} with the best-performing initial models as well as four self-alignment algorithms. 
 % \gd{also somewhere in this section, state something about what is computationally matched between approaches. I think it should be clear that Sparta involves more models, but that the number of inferences is the same...}
\begin{itemize}[leftmargin=*]
    \item \textbf{Best Initial Model (\textsc{Best Init})}, defined as $\arg \max_{M_i^0 \in \mathcal{M}^0} f(M_i^0)$, where $f(\cdot)$ represents the model's performance on the validation set of a specific task.
    \item \textbf{Self-Rewarding (\textsc{Self-reward})} \citep{yuan2025selfrewardinglanguagemodels}, where the model assigns rewards to its own generation via LLM-as-a-judge prompting and guides its own DPO training with self-generated preference data. 
    
    \item \textbf{Self-Meta-Rewarding (\textsc{Meta-reward})} \citep{wu2024metarewardinglanguagemodelsselfimproving}, 
    extends self-rewarding by having the model also evaluate its own reward judgments, leading to progressive improvement of both its response and reward generation abilities. 
        
    \item \textbf{SPIN} \citep{chen2024self}, a self-play-based fine-tuning method that iteratively aligns a weak LLM with the target data distribution by training it to distinguish its own previous generations from ground truth, enabling self-improvement without extra human annotation in later iterations. Notably, SPIN assumes access to the gold labels of training data, which is not required in \ourmethod{}. 
    
    \item \textbf{SPPO} \citep{wu2024selfplaypreferenceoptimizationlanguage}, where the model aligns itself through self-play, approximating the Nash equilibrium via iterative policy updates. Unlike purely unsupervised self-play, SPPO requires an \emph{external reward model} to perform annotation and generate win-rate signals, guiding policy refinement towards better distinguishing preferred from rejected responses. 

\end{itemize}
\vspace{-2pt}
Note that \textsc{Sparta} maintains a comparable or lower inference cost during training than other self-alignment methods, despite using more models, by processing each instruction only twice per iteration.
\paragraph{Dataset}  
We evaluate \ourmethod{} across 8 tasks and 12 datasets spanning three evaluation domains:
(1) Domain-Specific Question Answering, including MedQA-US (MedQA)~\citep{jin2021disease} and Normad~\citep{rao2024normad}. Normad is a benchmark assessing LLMs' cultural adaptability, comprising three subsets: country-based (Country), country-value-based (Value), and rule-of-thumb-based (RoT);
(2) Reasoning, covering GSM8K \citep{gsm8k}, Knowledge Crosswords (KCross) \citep{ding-etal-2024-knowledge}, COM$^2$ \citep{fang-etal-2024-complex}, and MATH \citep{hendrycksmeasuring}, where MATH is further divided into three difficulty levels: Easy (level 1 \& 2), Medium (level 3), and Hard (level 4 \& 5); (3) Instruction Following and Safety, evaluated on Alpaca \citep{dubois2024alpacafarm} for instruction-following and TruthfulQA (Truthful) \citep{lin2022truthfulqa} to for safety assessment. We report dataset statistics in Table~\ref{tab:dataset_stats} in the appendix.

\paragraph{Evaluation}  
We adopt task-specific evaluation metrics to assess model performance across diverse benchmarks. For tasks with discrete answer choices (Normad and KCross) and mathematical benchmarks (GSM8K, MATH), we report pass@1 accuracy. For open-ended generation tasks (Alpaca and COM$^2$), we use \emph{LLM-as-a-Judge}~\citep{zheng2023judging,bai2023benchmarking,dubois2024alpacafarm} approach and ask \textsc{Gemini‑1.5‑Flash}~\citep{team2023gemini} to assign score to model response $M(x)$ based on the task prompt $x$ and corresponding reference answer, on a 0-10 scale using a zero-shot prompting strategy.
For TruthfulQA, we follow the standard log-probability-based evaluation protocol, measuring whether the model assigns higher likelihood to truthful completions over false ones. Notably, we employ multi-faceted metrics in Appendix~\ref{metric-medqa} on MedQA to better align with real-world application requirements.

We execute \ourmethod{} in an unsupervised way over some seed instructions, then select the model in the pool with the highest performance on the validation set for each dataset, and report its performance on the held-out test set.

\section{Results}\label{results}
We present the performance of \textsc{Sparta} and baselines on 12 dataset settings in Table~\ref{tab:main}.

\textsc{Sparta} achieves state-of-the-art results in 10 out of 12 datasets. For reasoning and instruction-following tasks, \ourmethod{} outperforms the second-best baselines by 7\% on average. These results highlight the effectiveness of \ourmethod{} across a broad spectrum of tasks. We further dissect performance across evaluation categories.
\paravs
\paragraph{Domain-specific QA}
On the cultural adaptation benchmark Normad-Value subset, \ourmethod{} improves performance by 2\% compared to the best previous method \textsc{Self-reward}, highlighting the effectiveness of our collective alignment strategy. On MedQA, \ourmethod{} achieves the best performance in both clinical informativeness and logical consistency in Appendix~\ref{medqa_results}.
This indicates that \ourmethod{} promotes greater response diversity compared to self-alignment methods, enabling models to acquire broader knowledge, as well as more logical consistency across diverse domain-specific contexts. We present a detailed analysis of response diversity in Section~\ref{generation-diversity} and of preference diversity in Appendix~\ref{preference_diversity}.
\paravs
\paragraph{Reasoning}  
Across knowledge, mathematical and commonsense reasoning tasks, \textsc{Sparta} consistently delivers the strongest results. On GSM8K, \ourmethod{} improves by 4.5\% over the initialized model and 3.6\% over the best baseline \textsc{Sppo}.
On MATH (all levels), it shows consistent gains, averaging 4.0\% over the initialized model and 1.9\% over prior best. Furthermore, on COM$^2$, \ourmethod{} achieves 20.5\% improvement over initialization and 9.6\% over the strongest baseline \textsc{Self-reward}. These findings suggest that stochastic pairing effectively enhances multi-step reasoning and alignment across both arithmetic and commonsense domains.

\ourmethod{} provides more accurate reward signals by leveraging cross-model judgment, overcoming limitations common in self-alignment methods where self-feedback can amplify bias and overconfidence~\citep{xu2024prideprejudicellmamplifies,feuer2025styleoutweighssubstancefailure}. Additionally, we attribute the limited gains of \ourmethod{} on KCross to the small scale of the dataset (presented in Table~\ref{tab:dataset_stats}), where models may struggle to demonstrate robust generalization in the structural knowledge reasoning task with limited data. Nevertheless, these results indicate that with more unsupervised data available the alignment approach in \ourmethod{} will continue to improve with more diverse and extensive data.

\paravs~\paravs
\paragraph{Instruction Following and Truthfulness}
On the Alpaca benchmark for instruction following, \ourmethod{} outperforms the best initial model by 32.8\% and \textsc{Self-reward} by 28.1\%. Similarly, on TruthfulQA, \textsc{Sparta} (0.424) improves over the best initial model (0.410) and the previous best baseline from \textsc{Sppo} (0.421). 

The weaker performance of \textsc{Self-reward} and similar baselines arises from the inability of single-model feedback to fully capture the nuanced human-defined preferences and principles in instruction-following and safety tasks, while \ourmethod{} leverages multiple models to better encompass the range of human preferences.

Surprisingly, in \ourmethod{}, a model that initially underperforms has the potential to become the strongest in the model pool through iterative collaborative alignment. This phenomenon underscores the effectiveness of collective learning, where weaker models are not only retained but actively improved through diverse interactions with stronger peers. We present details in Table~\ref{tab:best-models}
\begin{table}[t]
\centering
\setlength{\tabcolsep}{4pt}
\renewcommand{\arraystretch}{1.1}
\centering
\setlength{\tabcolsep}{4pt}
\renewcommand{\arraystretch}{1.1}
\resizebox{1\textwidth}{!}{%
\begin{tabular}{l c ccc c c c ccc c c} 
\toprule[1.5pt]
\multirow{2}{*}{\textbf{Method}} & 
\multirow{2}{*}{\textbf{MedQA}} & 
\multicolumn{3}{c}{\textbf{Normad}} & 
\multirow{2}{*}{\textbf{KCross}} & 
\multirow{2}{*}{\textbf{GSM8K}} & 
\multirow{2}{*}{\textbf{COM$\textsuperscript{2}$}} & 
\multicolumn{3}{c}{\textbf{MATH}} & 
\multirow{2}{*}{\textbf{Alpaca}} & 
\multirow{2}{*}{\textbf{Truthful}} \\

\cmidrule(lr){3-5} \cmidrule(lr){9-11} 
& & \textbf{Country} & \textbf{Value} & \textbf{RoT} & 
& & & 
\textbf{Easy} & \textbf{Medium} & \textbf{Hard} & 
& \\ 
\midrule[0.75pt]

\textsc{Best Init} & .599 & .688 & .681 & .700 & .550 & .778 & 5.27 & \underline{.516} & \underline{.389} & .199 & 5.36 & .410 \\
\textsc{Self-reward} & \underline{.623} & \textbf{.699} & .692 & \textbf{.707} & .555 & .777 & \underline{5.74} & .513 & .376 & .188 & \underline{5.56} & .416 \\
\textsc{Meta-reward} & .618 & \underline{.692} & .680 & .700 & .550 & .779 & 5.47 & .503 & .385 & .202 & 5.49 & .413 \\
\textsc{SPIN} & .616 & .684 & .680 & .704 & \textbf{.580} & .782 & 5.58 & \underline{.516} & .369 & .204 & 5.49 & .420 \\
\textsc{SPPO} & .601 & .688 & \underline{.696} & .704 & .545 & \underline{.785} & 5.55 & .504 & .369 & \underline{.210} & \underline{5.56} & \underline{.421} \\
\midrule[0.75pt]
\textsc{Sparta} (ours) & \textbf{.662$^*$} & .688 & \textbf{.715} & \textbf{.707} & \underline{.560} & \textbf{.813$^*$} & \textbf{6.35$^*$} & \textbf{.530} & \textbf{.396} & \textbf{.212} & \textbf{7.12$^*$} & \textbf{.424} \\
\bottomrule[1.5pt]
\end{tabular}
}
\vspace*{5pt}
\caption{
Performance of baselines and \ourmethod{} across 12 datasets. Best results are in \textbf{bold}, second-best are \underline{underlined}. Different-task results of \ourmethod{} with * are significantly better than best baselines at the level of $0.05$.
}
\vspace{-20pt}
\label{tab:main}
\end{table}
\secvsabove
\section{Analysis}
\secvsbelow
\paragraph{Generalization}
We assess \ourmethod{}'s generalization capability using MATH~\citep{hendrycksmeasuring}—a benchmark designed to evaluate LLM's reasoning ability across different levels of mathematical complexity, including Easy, Medium, and Hard~\citep{sun2024easytohard}.

\begin{wrapfigure}{h}{0.53\textwidth}
\centering 
\vspace{-5pt}
\includegraphics[width=\linewidth]{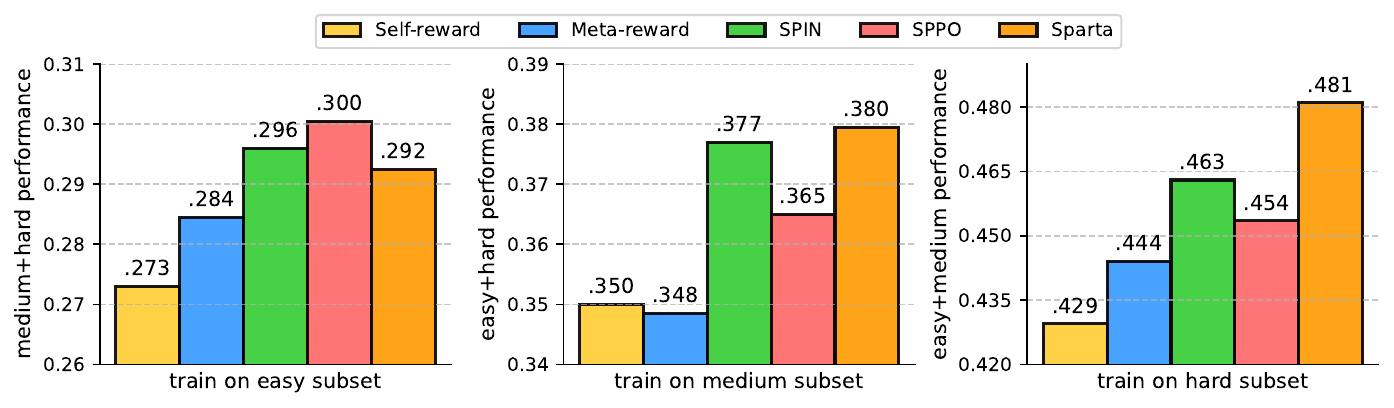}
\caption{Cross-subset generalization accuracy on the MATH benchmark. Each group of bars corresponds to a training subset (Easy, Medium, Hard), with OOD performance measured on the two held-out subsets.}
\label{fig:generalization}
\end{wrapfigure}

As shown in Figure~\ref{fig:generalization}, we evaluate \ourmethod{} by fine-tuning a pool of models on each subset individually and testing their performance on the other two subsets with different difficulty levels. 
% This experimental design assesses the model's cross-difficulty generalization ability, probing its capacity to transfer mathematical reasoning from one complexity level to another. 
Results in
% \gd{break figure into 3 subfigures so different test sets don't share a y-axis; this is a bit confusing} 
Figure~\ref{fig:generalization} show that \ourmethod{} consistently outperforms all baselines when trained on Medium and Hard subsets, demonstrating \ourmethod{}'s superior capability in generalizing across mathematical complexities.
% , which is attributed to two key design principles. First, the reputation-based optimization encourages models to solve problems through logical reasoning and step-wise validation, promoting robust generalization across difficulty gradients. This optimization\gd{These reasons feel a bit made up; they are a bit vague and I don't have intuition (or evidence) for Sparta doing these things. I would prefer to just say: we tested on OOD generalization and found it's good} strategy rewards models that consistently succeed across diverse mathematical challenges, leading to stronger adaptability to unseen problem structures. Second, the stochastic match-making mechanism in \ourmethod{} exposes models to a variety of problem-solving paths during training, increasing exposure to non-trivial solutions and complex reasoning processes. This enhanced exposure to mathematical diversity allows \ourmethod{} to develop more abstract problem-solving strategies, enabling stronger transferability across varying levels of mathematical complexity. Together, these mechanisms equip \ourmethod{} with the ability to bridge the gap between structurally distinct yet conceptually related mathematical tasks.
\paravs
\paragraph{Scaling the Model Pool}

\begin{wrapfigure}{r}{0.5\textwidth}
    \vspace{-15pt}
    \centering
    \includegraphics[width=\linewidth]{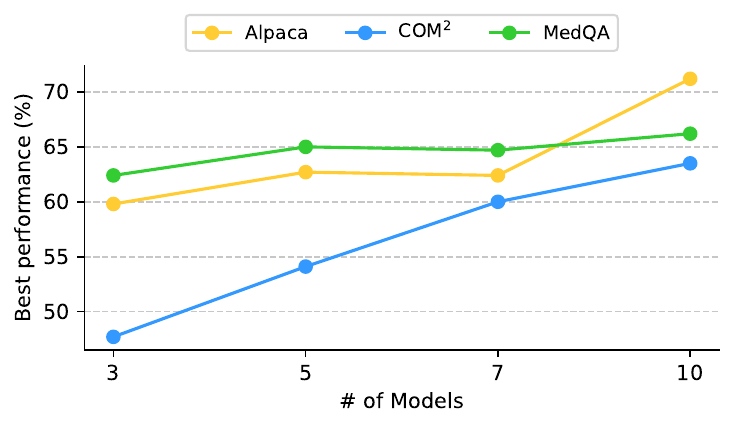}
    \caption{
    Effect of pool size on alignment performance. We vary the number of candidate models participating in each training round and measure the final performance. Results show that larger pools lead to better outcomes, indicating that \ourmethod{} benefits from having more diverse LLMs as participants.
    }
    \label{fig:pool_size}
    \vspace*{-15pt}
\end{wrapfigure}
\ourmethod{} utilizes a larger pool of models to encourage diverse outputs and enable cross-model judgment. This naturally raises the question: \textit{How does the number of models affect the quality of the best aligned model in the system?}
To investigate this, we conduct a series of experiments (details in Appendix~\ref{scaling}). As shown in Figure~\ref{fig:pool_size}, increasing the number of models in the alignment pool—by selecting a larger k for the top-k models from an initial set of 10-consistently improves performance across all four benchmarks.
% \gd{say something here about how it's always the best $k$ models}
Compared to the 3-model setting, the 10-model pool yields a performance increase of 19.1\% on Alpaca, 33.1\% on COM$^2$ and 6.1\% on MedQA. These results suggest a positive benefit in larger model pool sizes, where more models provide richer diversity and stronger supervision, leading to better aligned models.

% \paravs
\paragraph{Generation Diversity}\label{generation-diversity}
\begin{wraptable}{r}{0.5\textwidth}
\scriptsize
\centering
\setlength{\tabcolsep}{3pt}
\renewcommand{\arraystretch}{1.1}
\vspace*{-14pt}
\resizebox{0.5\textwidth}{!}{
\begin{tabular}{lcccc}
\toprule[1pt]
\textbf{Method} & \textbf{Edit} & \textbf{BLEU} & \textbf{Embed} & \textbf{LLM} \\
\midrule[0.75pt]
\multicolumn{5}{c}{\textbf{Baselines (1 Model)}} \\
\textsc{Self-reward (Init)} & 0.667 & 0.859 & 0.129 & 5.73 \\
\textsc{Self-reward (Final)} & 0.673 & 0.867 & 0.134 & 5.90 \\
\addlinespace[0.3em]
\hdashline
\addlinespace[0.3em]
\textsc{Meta-reward (Init)} & 0.653 & 0.876 & 0.120 & 5.99 \\
\textsc{Meta-reward (Final)} & 0.655 & 0.878 & 0.122 & 5.93 \\

\midrule[0.5pt]
\rowcolor{lightgray}
\multicolumn{5}{c}{\textbf{Sparta (10 Models)}} \\
\textsc{Sparta (Init)} & \textbf{0.732} & \textbf{0.946} & \textbf{0.178} & \underline{6.12} \\
\textsc{Sparta (Final)} & \underline{0.722} & \underline{0.918} & \underline{0.142} & \textbf{6.42} \\
\bottomrule[1pt]
\end{tabular}
}
\vspace*{-5pt}
\caption{
Output diversity across responses after the first iteration and the final iteration on the instruction-following task, using automatic and LLM-based metrics. \textsc{Sparta} employs 10 models compared to 1 model for baselines. \textsc{Sparta} outperforms the two baselines across all dimensions.
}
\label{tab:inst-metrics}
\vspace*{-10pt}
\end{wraptable}
\vspace*{-5pt}
To test the hypothesis that multi-LLM collaboration produces more diverse outputs than a single model—thereby offering richer and more robust learning data—we conduct controlled experiments comparing models trained with \textsc{Self-reward}
and \textsc{Meta-reward} versus \ourmethod{}. To align with previous experiment settings, models in \textsc{Self-reward} and \textsc{Meta-reward}. We evaluate the first and final models from \textsc{Self-reward} and \textsc{Meta-reward} and the corresponding models from \ourmethod{}. To align with our main experiments, in \textsc{Self-reward} and \textsc{Meta-reward}, we use a single model to generate 10 responses to a prompt, while in \ourmethod{} we use 10 models to output separately. Response pairwise diversity is assessed using four complementary metrics in Appendix~\ref{evaluation-metric}, comprising two traditional metrics and two LLM-based metrics.
As shown in Table~\ref{tab:inst-metrics}, \ourmethod{} outperforms \textsc{Self-reward} and \textsc{Meta-reward} across all metrics, confirming that multi-model collaboration introduces greater output diversity across lexical, structural, and semantic dimensions, further supporting more effective alignment. Figure~\ref{fig:trade-off} shows a trade-off in our system that the output diversity decreases as the iteration processes, while the best performance increases. With MedQA~\citep{jin2021disease} as an example, we analyze the underlying mechanism transferring generation diversity to  improvement~\citep{kirk2023understanding}. 

Table~\ref{tab:entropy-metrics} and Appendix~\ref{generation-diversity} reveal that, compared to \textsc{Self-reward}, preference pairs in \ourmethod{} exhibit greater diversity between the chosen and rejected samples from the angles of semantic similarity and length. This increased variance facilitates more effective learning of latent patterns through preference optimization~\citep{dpo,naik2024diversity,slocum2025diverse}. Furthermore, the selected samples are consistently more complete, clinically informative, direct, and logically structured than their rejected counterparts, contributing to enhanced model alignment and robustness.

\paragraph{Model Diversity Matters}
\begin{wrapfigure}{r}{0.5\textwidth}
     \vspace{-10pt}
     \centering
     \includegraphics[width=\linewidth]{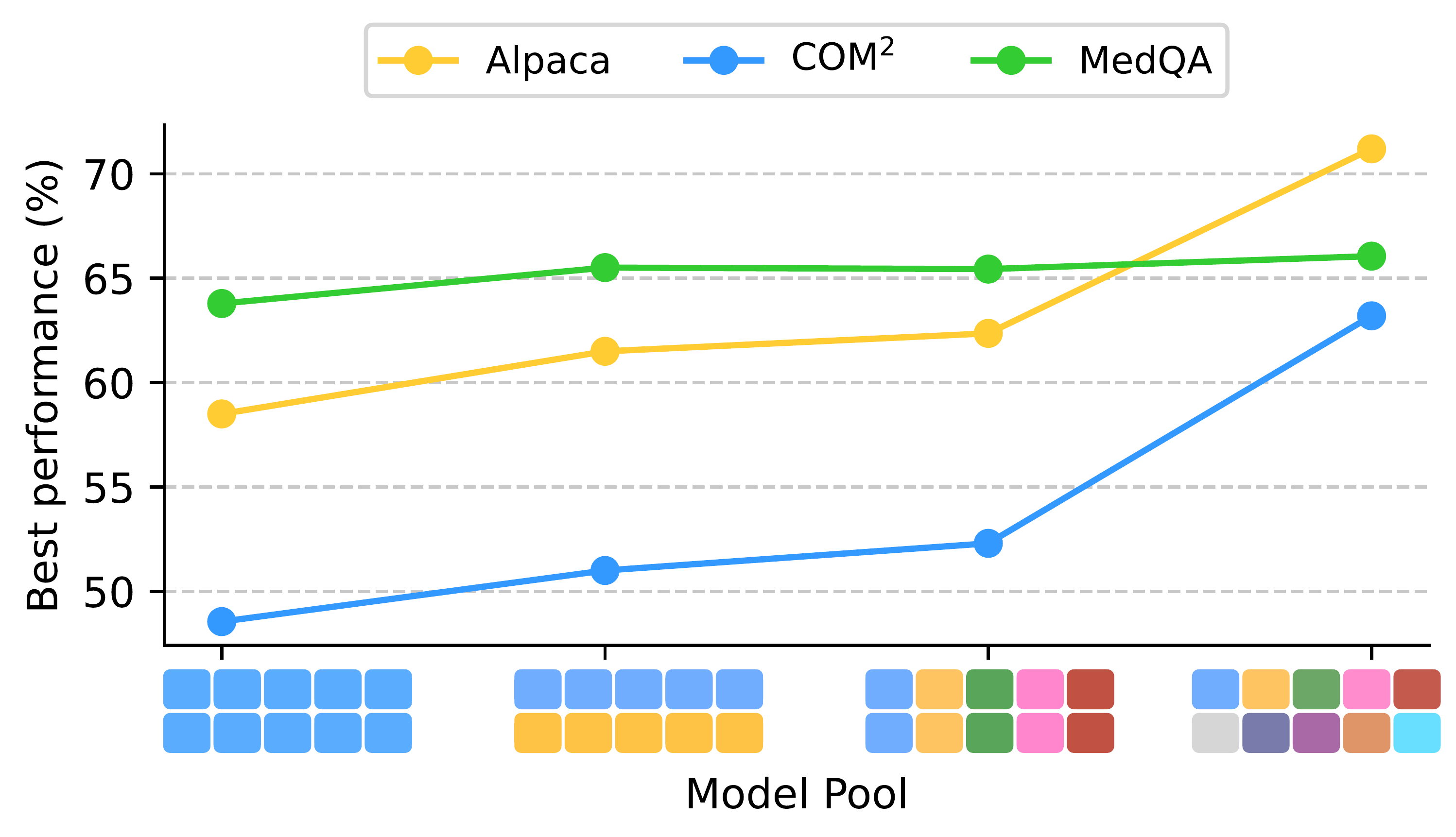}
     \caption{
     Impact of model pool diversity on alignment performance. The x-axis shows the configurations of model pool diversity: $1 \times10, 2\times5, 5\times2 \text{ and } 10\times1$. The results demonstrate that increasing the diversity of the model pool consistently improves performance.
     }
     \vspace{-15pt}
     \label{fig:model_diversity}
\end{wrapfigure}
In addition to pool size, we ask: \textit{How does the diversity of the model pool affect the final performance?}

To answer this, we conduct experiments on three representative datasets—Alpaca, COM$^2$ and MedQA. For each dataset, we vary the diversity of the initial model pool by sampling $a$ distinct LLMs,
% \gd{make it clear that you always have best initial model in there} 
each repeated $b$ times, and use the resulting $a \times b$ pool as the starting point for \ourmethod{}. For fairness, we always keep the best initial model in the pool. In Figure~\ref{fig:model_diversity}, we report results for configurations $1\times 10, 2\times5, 5\times2, \text{ and } 10\times1$. As shown in Figure~\ref{fig:model_diversity}, performance consistently improves with greater model diversity. Notably, the fully diverse setting $10\times1$ outperforms the fully repetitive setting $1\times10$ by 18.5\% averaged across the three datasets. These results provide strong empirical evidence that \textit{Model Diversity Matters}, that \textsc{Sparta} benefits from multiple models with heterogeneous skills complementing each other in the collective alignment process.

\paragraph{Correlation between Reputation and Performance}
\begin{wrapfigure}{r}{0.5\textwidth}
     \centering
     \vspace{-15pt}
     \includegraphics[width=\linewidth]{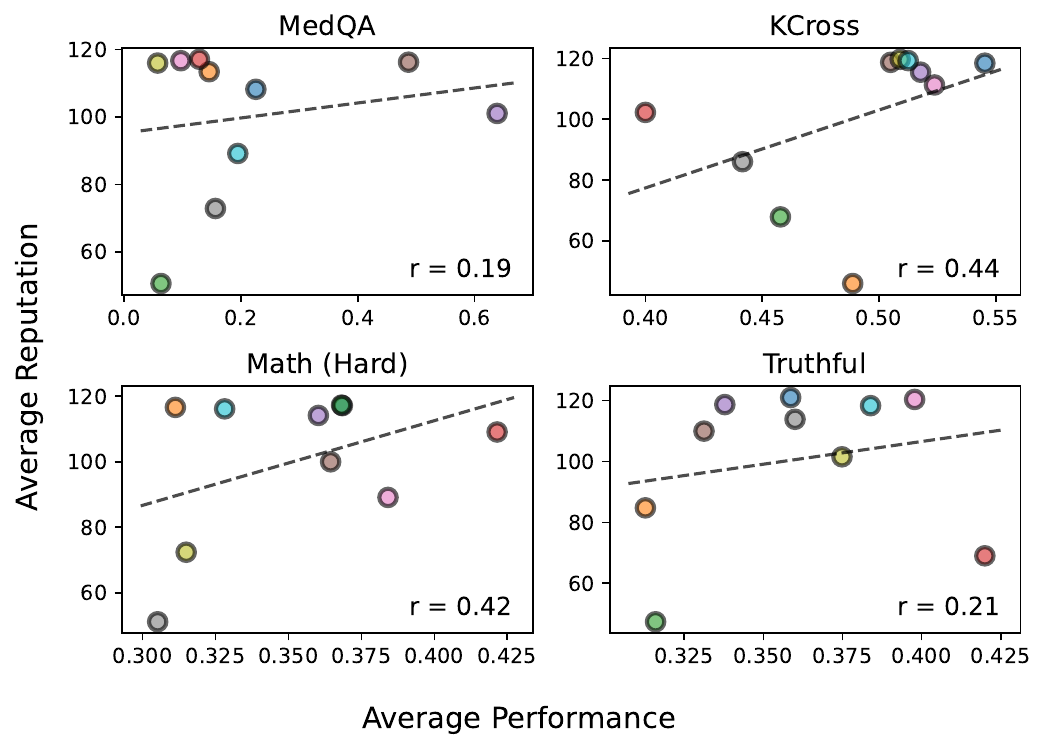}
     \caption{
     Correlation between a model's average performance on a specific task and its average reputation in the model pool. The 10 points in each subplot indicate 10 models. $r$ stands for Pearson correlation coefficient.
     }
     \vspace{-13pt}
     \label{fig:correlation}
\end{wrapfigure}
To investigate how well the peer-derived reputation reflects actual model performance, we analyze the correlation between models' average performance and their corresponding reputation scores over iterations under \ourmethod{}. 
Figure~\ref{fig:correlation} reveals a positive correlation between reputation scores and performance, which suggests that the reputation mechanism effectively captures meaningful signals about model alignment, providing empirical validation for our approach. We present Figure~\ref{fig:correlation_all_r} to provide a comprehensive view of the correlation across all evaluated datasets, reporting an average correlation coefficient of 0.21 across all tasks.
Furthermore, Figure~\ref{fig:alpaca_80} shows that \ourmethod{} induces a hierarchical stratification among the 10 models, hinting at potential top-down influence.

\paragraph{Ablation Study} 
\begin{wraptable}{r}{0.5\textwidth}
\scriptsize
\centering
\setlength{\tabcolsep}{3pt}
\renewcommand{\arraystretch}{1.1}
\vspace*{-10pt}
\resizebox{0.5\textwidth}{!}{
\begin{tabular}{lcccc}
\toprule[1pt]
\textbf{Method} & \textbf{Alpaca} & \textbf{COM$^2$} & \textbf{MedQA} & \textbf{GSM8K} \\
\midrule[0.75pt]
\textsc{Sparta} & \textbf{7.12} & \textbf{6.35} & \textbf{0.662} & \textbf{0.813} \\
- w/o reputation updates & 6.17 & 5.49 & 0.647 & 0.765 \\
- w/o downweighting & 6.63 & 6.10 & 0.651 & 0.781 \\
- w/o randomness in match & 6.04 & 5.73 & 0.654 & 0.779 \\
- w/o top-$k$ match & 6.74 & 5.96 & 0.659 & 0.804 \\
\bottomrule[1pt]
\end{tabular}
}
\vspace*{-5pt}
\caption{
Ablation of \ourmethod{} across four datasets. All components jointly contribute to the best performance across all tasks.
}
\label{tab:ablation}
\vspace*{-10pt}
\end{wraptable}
To evaluate the importance of individual components in \ourmethod{}, we conduct an ablation study on four datasets: Alpaca, COM$^2$, MedQA, and GSM8K. As shown in Table~\ref{tab:ablation}, removing any single component leads to consistent performance drop, confirming that each design choice contributes critically to the system's overall effectiveness.

The removal of randomness in matchmaking (\textit{w/o randomness in match}) degrades performance most severely on Alpaca and COM$^2$, suggesting that stochastic opponent selection encourages behavioral diversity and improves generalization in instruction-following settings. Finally, while omitting the top-$k$ matching constraint (\textit{w/o top-$k$ match}) yields marginal improvements on certain datasets, it compromises stability on others, indicating that constrained pairing is crucial for balancing exploration and reliable supervision. These results demonstrate that the strong performance of \ourmethod{} arises from the synergy of all components, and that omitting any part leads to measurable and often substantial regressions.

\secvsabove
\section{Related Work}
\secvsbelow
\paragraph{Reinforcement Learning from AI Feedback (RLAIF)}
Aligning large language models (LLMs) with human preferences has been a cornerstone of their success~\citep{leike2018scalable,ziegler2019fine,ouyang2022training}. Traditional methods like RLHF involve collecting human-generated preference labels with reinforcement learning to distinguish responses~\citep{schulman2017proximal,dpo}, which can be costly and time-consuming~\citep{ouyang2022training,bai2022training}. This limitation has spurred interest in alternative approaches, such as RLAIF, where AI-generated feedback is used to train reward models~\citep{ouyang2022training,anwar2024foundational,yuan2025selfrewardinglanguagemodels,fisch2024robust, schulman2017proximal,jiang2023llm, eisenstein2025don}. Constitutional AI~\citep{bai2022constitutionalaiharmlessnessai}  uses a set of predefined principles to guide an LLM in critiquing and revising its responses to harmful queries. These revisions are then used to fine-tune the AI model through reinforcement learning with AI-generated feedback. Self-Rewarding~\citep{yuan2025selfrewardinglanguagemodels} leverages an initial model to generate multiple responses, rank them using LLM-as-a-Judge~\citep{zheng2023judging,bai2023benchmarking,dubois2024alpacafarm}, and perform preference learning. Meta-Rewarding~\citep{wu2024metarewardinglanguagemodelsselfimproving} allows the model to assess and iteratively refine its own responses and reward generation. Other self-alignment methods~\citep{wu2024selfplaypreferenceoptimizationlanguage, chen2024self, huang2022large, zelikman2022star, chen2024self, guo2024direct, zhou2024calibrated,matrix_icml2024} either depend on ground truth or require signals from a larger, more capable model. Despite their potential, previous work suggests that self-bias from self-feedback~\citep{xu2024prideprejudicellmamplifies,feuer2025styleoutweighssubstancefailure} is a key bottleneck preventing further self-improvement. Additionally, the decreasing diversity resulting from alignment causes the distinction between positive and negative responses to become increasingly blurred~\citep{kirk2023understanding,wang2022adamixmixtureofadaptationsparameterefficientmodel,xiao2024algorithmicbiasaligninglarge,wang2023reverseklgeneralizingdirect,slocum2025diverse}, leading to preference inconsistency during self-alignment~\citep{wang2025creamconsistencyregularizedselfrewarding, reward_inconsistency, zhou2025selfconsistencyinternalrewardmodels}, which in turn negatively impacts alignment effectiveness. In contrast, \ourmethod{} mitigates these effects by introducing multiple diverse LLMs to generate responses and cross-judge them.
\paravs
\paragraph{Multi-LLM Collaboration}
While pushing a single general-purpose LLM to achieve new benchmarks is valuable, an increasing body of research seeks to advance beyond a single model and explore model collaboration~\citep{feng2025llmdroolsmultillmcollaboration, feng2024modular, subramaniam2025multiagent}. This is due to three main limitations: (1) A single LLM underrepresents real-world linguistic diversity, evolving trends, and domain-specific knowledge~\citep{lazaridou2021pitfalls,dhingra2022time, kasai2024realtime}; (2) No single LLM excels across all tasks, highlighting the need for model specialization~\citep{song2023globalbench,faisal-etal-2024-dialectbench,wan2024dellgeneratingreactionsexplanations}; (3) A single LLM fails to represent the diverse needs, values, and socio-cultural backgrounds of its users~\citep{sorensenposition, feng2024modular, leibo2024theory, santurkar2023whose, feng2023pretraining}.

Therefore, researchers are probing the potential of \emph{multi-LLM collaboration}~\citep{feng2025llmdroolsmultillmcollaboration}, by allowing LLMs to evaluate and reflect on each other's outputs~\citep{cohen2023lm}, offers a promising approach to improving factual accuracy~\citep{jiang2024tigerscore,factscore,feng2023factkbgeneralizablefactualityevaluation} and reducing hallucinations~\citep{feng-etal-2024-dont,feng2025teachingllmsabstainlanguages}, addressing the limitations of single-model self-reflection~\citep{wang2022self, xu2024sayself, shinn2024reflexion, madaan2024self} and confirmation biases~\citep{ji2023survey}.
Prior studies~\citep{xiong2023examining, liang2023encouraging, du2024improving} suggest that multi-LLM systems can improve their factuality through multi-turn debates~\citep{du2024improving}.
Furthermore, multi-LLM collaboration, through modular systems~\citep{feng2024modular}, multi-LLM as a judge \citep{zhao2024language} and diverse reward models~\citep{jang2023personalized, tao2025weakllmsecretlystrong}, offers a solution to address the cultural, political, and social biases inherent in LLMs, promoting fairness and pluralism~\citep{feng2024modular,fisher2025politicalneutralityaiimpossible} in alignment. \ourmethod{} leverages multi-LLM collaboration and competition and goes beyond self-alignment, outperforming single-model methods on instruction-following~\citep{dubois2024alpacafarm}, reasoning~\citep{gsm8k,chao2024match,ding-etal-2024-knowledge,fang-etal-2024-complex}, cultural adaptation tasks~\citep{rao2024normad}, and more.
\section{Conclusion}
We propose \ourmethod{}, a game-theoretic algorithm to collectively improve multiple
LLMs through competition. 
Starting with a pool of diverse models and a dataset, we iteratively select pairs of models to combat by generating responses to sampled prompts from the dataset.
The remaining models in the pool act as judges to evaluate the responses. A reliable reward signal is obtained by aggregating their judgments, weighted by each judge's reputation. 
Winners of each combat see their reputation increase, while losers experience a corresponding decrease. 
Extensive experiments show that \ourmethod{} outperforms the best baseline by up to 28.1\%, with an average performance increase of 7\% across 10 tasks spanning several domains. Further analysis reveals that \ourmethod{} better generalizes to unseen data settings, model reputation scores are positively correlated with task performance, and that \ourmethod{} benefits from a larger and diverse pool of participants.

\section*{Acknowledgments}
This research was developed with funding from the Defense Advanced Research Projects Agency's (DARPA) SciFy program (Agreement No. HR00112520300). The views expressed are those of the author and do not reflect the official policy or position of the Department of Defense or the U.S.~Government.
This material is based upon work supported by the Defense Advanced Research Projects Agency and the Air Force Research Laboratory, contract number(s): FA8650-23-C-7316. Any opinions, findings and conclusions, or recommendations expressed in this material are those of the author(s) and do not necessarily reflect the views of AFRL or DARPA.
We gratefully acknowledge gift and compute funding from Amazon and Google.
 Shangbin Feng would like to thank the support of the IBM PhD Fellowship and the Jane Street Graduate Research Fellowship.

\bibliography{neurips_2025}

\begin{thebibliography}{112}
\providecommand{\natexlab}[1]{#1}
\providecommand{\url}[1]{\texttt{#1}}
\expandafter\ifx\csname urlstyle\endcsname\relax
  \providecommand{\doi}[1]{doi: #1}\else
  \providecommand{\doi}{doi: \begingroup \urlstyle{rm}\Url}\fi

\bibitem[Anwar et~al.(2024)Anwar, Saparov, Rando, Paleka, Turpin, Hase, Lubana, Jenner, Casper, Sourbut, et~al.]{anwar2024foundational}
Usman Anwar, Abulhair Saparov, Javier Rando, Daniel Paleka, Miles Turpin, Peter Hase, Ekdeep~Singh Lubana, Erik Jenner, Stephen Casper, Oliver Sourbut, et~al.
\newblock Foundational challenges in assuring alignment and safety of large language models.
\newblock \emph{arXiv preprint arXiv:2404.09932}, 2024.

\bibitem[Bai et~al.(2022{\natexlab{a}})Bai, Jones, Ndousse, Askell, Chen, DasSarma, Drain, Fort, Ganguli, Henighan, et~al.]{bai2022training}
Yuntao Bai, Andy Jones, Kamal Ndousse, Amanda Askell, Anna Chen, Nova DasSarma, Dawn Drain, Stanislav Fort, Deep Ganguli, Tom Henighan, et~al.
\newblock Training a helpful and harmless assistant with reinforcement learning from human feedback.
\newblock \emph{arXiv preprint arXiv:2204.05862}, 2022{\natexlab{a}}.

\bibitem[Bai et~al.(2022{\natexlab{b}})Bai, Kadavath, Kundu, Askell, Kernion, Jones, Chen, Goldie, Mirhoseini, McKinnon, Chen, Olsson, Olah, Hernandez, Drain, Ganguli, Li, Tran-Johnson, Perez, Kerr, Mueller, Ladish, Landau, Ndousse, Lukosuite, Lovitt, Sellitto, Elhage, Schiefer, Mercado, DasSarma, Lasenby, Larson, Ringer, Johnston, Kravec, Showk, Fort, Lanham, Telleen-Lawton, Conerly, Henighan, Hume, Bowman, Hatfield-Dodds, Mann, Amodei, Joseph, McCandlish, Brown, and Kaplan]{bai2022constitutionalaiharmlessnessai}
Yuntao Bai, Saurav Kadavath, Sandipan Kundu, Amanda Askell, Jackson Kernion, Andy Jones, Anna Chen, Anna Goldie, Azalia Mirhoseini, Cameron McKinnon, Carol Chen, Catherine Olsson, Christopher Olah, Danny Hernandez, Dawn Drain, Deep Ganguli, Dustin Li, Eli Tran-Johnson, Ethan Perez, Jamie Kerr, Jared Mueller, Jeffrey Ladish, Joshua Landau, Kamal Ndousse, Kamile Lukosuite, Liane Lovitt, Michael Sellitto, Nelson Elhage, Nicholas Schiefer, Noemi Mercado, Nova DasSarma, Robert Lasenby, Robin Larson, Sam Ringer, Scott Johnston, Shauna Kravec, Sheer~El Showk, Stanislav Fort, Tamera Lanham, Timothy Telleen-Lawton, Tom Conerly, Tom Henighan, Tristan Hume, Samuel~R. Bowman, Zac Hatfield-Dodds, Ben Mann, Dario Amodei, Nicholas Joseph, Sam McCandlish, Tom Brown, and Jared Kaplan.
\newblock Constitutional ai: Harmlessness from ai feedback, 2022{\natexlab{b}}.
\newblock URL \url{https://arxiv.org/abs/2212.08073}.

\bibitem[Bai et~al.(2023)Bai, Ying, Cao, Lv, He, Wang, Yu, Zeng, Xiao, Lyu, Zhang, Li, and Hou]{bai2023benchmarking}
Yushi Bai, Jiahao Ying, Yixin Cao, Xin Lv, Yuze He, Xiaozhi Wang, Jifan Yu, Kaisheng Zeng, Yijia Xiao, Haozhe Lyu, Jiayin Zhang, Juanzi Li, and Lei Hou.
\newblock Benchmarking foundation models with language-model-as-an-examiner.
\newblock In \emph{Thirty-seventh Conference on Neural Information Processing Systems Datasets and Benchmarks Track}, 2023.

\bibitem[Chao et~al.(2024)Chao, Zhao, Jiao, Li, Liu, and Yang]{chao2024match}
Wang Chao, Jiaxuan Zhao, Licheng Jiao, Lingling Li, Fang Liu, and Shuyuan Yang.
\newblock A match made in consistency heaven: when large language models meet evolutionary algorithms.
\newblock \emph{arXiv preprint arXiv:2401.10510}, 2024.

\bibitem[Chaudhary(2023)]{codealpaca}
Sahil Chaudhary.
\newblock Code alpaca: An instruction-following llama model for code generation.
\newblock \url{https://github.com/sahil280114/codealpaca}, 2023.

\bibitem[Chen et~al.(2025)Chen, Chen, Rege, Wang, and Vinayak]{chen2025pal}
Daiwei Chen, Yi~Chen, Aniket Rege, Zhi Wang, and Ramya~Korlakai Vinayak.
\newblock {PAL}: Sample-efficient personalized reward modeling for pluralistic alignment.
\newblock In \emph{The Thirteenth International Conference on Learning Representations}, 2025.
\newblock URL \url{https://openreview.net/forum?id=1kFDrYCuSu}.

\bibitem[Chen et~al.(2024{\natexlab{a}})Chen, Prasad, Saha, Stengel-Eskin, and Bansal]{chih2024magicore}
Justin Chih-Yao Chen, Archiki Prasad, Swarnadeep Saha, Elias Stengel-Eskin, and Mohit Bansal.
\newblock Magicore: Multi-agent, iterative, coarse-to-fine refinement for reasoning.
\newblock \emph{arXiv e-prints}, pages arXiv--2409, 2024{\natexlab{a}}.

\bibitem[Chen et~al.(2024{\natexlab{b}})Chen, Deng, Yuan, Ji, and Gu]{chen2024self}
Zixiang Chen, Yihe Deng, Huizhuo Yuan, Kaixuan Ji, and Quanquan Gu.
\newblock Self-play fine-tuning converts weak language models to strong language models.
\newblock \emph{arXiv preprint arXiv:2401.01335}, 2024{\natexlab{b}}.

\bibitem[Christiano et~al.(2017)Christiano, Leike, Brown, Martic, Legg, and Amodei]{christiano2017}
Paul~F. Christiano, Jan Leike, Tom~B. Brown, Miljan Martic, Shane Legg, and Dario Amodei.
\newblock Deep reinforcement learning from human preferences.
\newblock In \emph{Advances in Neural Information Processing Systems 30: Annual Conference on Neural Information Processing Systems 2017, December 4-9, 2017, Long Beach, CA, {USA}}, pages 4299--4307, 2017.

\bibitem[Chung et~al.(2024)Chung, Hou, Longpre, Zoph, Tay, Fedus, Li, Wang, Dehghani, Brahma, et~al.]{chung2024scaling}
Hyung~Won Chung, Le~Hou, Shayne Longpre, Barret Zoph, Yi~Tay, William Fedus, Yunxuan Li, Xuezhi Wang, Mostafa Dehghani, Siddhartha Brahma, et~al.
\newblock Scaling instruction-finetuned language models.
\newblock \emph{Journal of Machine Learning Research}, 25\penalty0 (70):\penalty0 1--53, 2024.

\bibitem[Cobbe et~al.(2021)Cobbe, Kosaraju, Bavarian, Chen, Jun, Kaiser, Plappert, Tworek, Hilton, Nakano, Hesse, and Schulman]{gsm8k}
Karl Cobbe, Vineet Kosaraju, Mohammad Bavarian, Mark Chen, Heewoo Jun, Lukasz Kaiser, Matthias Plappert, Jerry Tworek, Jacob Hilton, Reiichiro Nakano, Christopher Hesse, and John Schulman.
\newblock Training verifiers to solve math word problems.
\newblock \emph{CoRR}, abs/2110.14168, 2021.

\bibitem[Cohen et~al.(2023)Cohen, Hamri, Geva, and Globerson]{cohen2023lm}
Roi Cohen, May Hamri, Mor Geva, and Amir Globerson.
\newblock Lm vs lm: Detecting factual errors via cross examination.
\newblock \emph{arXiv preprint arXiv:2305.13281}, 2023.

\bibitem[Cui et~al.(2025)Cui, Yuan, Wang, Wang, Li, He, Fan, Yu, Xu, Chen, Yuan, Chen, Zhang, Lv, Wang, Yao, Han, Peng, Cheng, Liu, Sun, Zhou, and Ding]{cui2025processreinforcementimplicitrewards}
Ganqu Cui, Lifan Yuan, Zefan Wang, Hanbin Wang, Wendi Li, Bingxiang He, Yuchen Fan, Tianyu Yu, Qixin Xu, Weize Chen, Jiarui Yuan, Huayu Chen, Kaiyan Zhang, Xingtai Lv, Shuo Wang, Yuan Yao, Xu~Han, Hao Peng, Yu~Cheng, Zhiyuan Liu, Maosong Sun, Bowen Zhou, and Ning Ding.
\newblock Process reinforcement through implicit rewards, 2025.
\newblock URL \url{https://arxiv.org/abs/2502.01456}.

\bibitem[Dhingra et~al.(2022)Dhingra, Cole, Eisenschlos, Gillick, Eisenstein, and Cohen]{dhingra2022time}
Bhuwan Dhingra, Jeremy~R Cole, Julian~Martin Eisenschlos, Daniel Gillick, Jacob Eisenstein, and William~W Cohen.
\newblock Time-aware language models as temporal knowledge bases.
\newblock \emph{Transactions of the Association for Computational Linguistics}, 10:\penalty0 257--273, 2022.

\bibitem[Ding et~al.(2024)Ding, Feng, Liu, Tan, Balachandran, He, and Tsvetkov]{ding-etal-2024-knowledge}
Wenxuan Ding, Shangbin Feng, Yuhan Liu, Zhaoxuan Tan, Vidhisha Balachandran, Tianxing He, and Yulia Tsvetkov.
\newblock Knowledge crosswords: Geometric knowledge reasoning with large language models.
\newblock In \emph{Findings of the Association for Computational Linguistics ACL 2024}, 2024.

\bibitem[Du et~al.(2024)Du, Li, Torralba, Tenenbaum, and Mordatch]{du2024improving}
Yilun Du, Shuang Li, Antonio Torralba, Joshua~B. Tenenbaum, and Igor Mordatch.
\newblock Improving factuality and reasoning in language models through multiagent debate.
\newblock In \emph{Forty-first International Conference on Machine Learning}, 2024.
\newblock URL \url{https://openreview.net/forum?id=zj7YuTE4t8}.

\bibitem[Dubois et~al.(2024)Dubois, Li, Taori, Zhang, Gulrajani, Ba, Guestrin, Liang, and Hashimoto]{dubois2024alpacafarm}
Yann Dubois, Chen~Xuechen Li, Rohan Taori, Tianyi Zhang, Ishaan Gulrajani, Jimmy Ba, Carlos Guestrin, Percy~S Liang, and Tatsunori~B Hashimoto.
\newblock Alpacafarm: A simulation framework for methods that learn from human feedback.
\newblock \emph{Advances in Neural Information Processing Systems}, 36, 2024.

\bibitem[Ebtekar and Liu(2021)]{ebtekar2021elo}
Aram Ebtekar and Paul Liu.
\newblock Elo-mmr: A rating system for massive multiplayer competitions.
\newblock In \emph{Proceedings of the Web Conference 2021}, pages 1772--1784, 2021.

\bibitem[Eisenstein et~al.(2025)Eisenstein, Aghajani, Fisch, Dua, Huot, Lapata, Zayats, and Berant]{eisenstein2025don}
Jacob Eisenstein, Reza Aghajani, Adam Fisch, Dheeru Dua, Fantine Huot, Mirella Lapata, Vicky Zayats, and Jonathan Berant.
\newblock Don't lie to your friends: Learning what you know from collaborative self-play.
\newblock \emph{arXiv preprint arXiv:2503.14481}, 2025.

\bibitem[Faisal et~al.(2024)Faisal, Ahia, Srivastava, Ahuja, Chiang, Tsvetkov, and Anastasopoulos]{faisal-etal-2024-dialectbench}
Fahim Faisal, Orevaoghene Ahia, Aarohi Srivastava, Kabir Ahuja, David Chiang, Yulia Tsvetkov, and Antonios Anastasopoulos.
\newblock {DIALECTBENCH}: An {NLP} benchmark for dialects, varieties, and closely-related languages.
\newblock In Lun-Wei Ku, Andre Martins, and Vivek Srikumar, editors, \emph{Proceedings of the 62nd Annual Meeting of the Association for Computational Linguistics (Volume 1: Long Papers)}, 2024.

\bibitem[Fang et~al.(2024)Fang, Chen, Song, and Bosselut]{fang-etal-2024-complex}
Tianqing Fang, Zeming Chen, Yangqiu Song, and Antoine Bosselut.
\newblock Complex reasoning over logical queries on commonsense knowledge graphs.
\newblock In Lun-Wei Ku, Andre Martins, and Vivek Srikumar, editors, \emph{Proceedings of the 62nd Annual Meeting of the Association for Computational Linguistics (Volume 1: Long Papers)}, pages 11365--11384, Bangkok, Thailand, August 2024. Association for Computational Linguistics.
\newblock \doi{10.18653/v1/2024.acl-long.613}.
\newblock URL \url{https://aclanthology.org/2024.acl-long.613/}.

\bibitem[Feng et~al.(2023{\natexlab{a}})Feng, Balachandran, Bai, and Tsvetkov]{feng2023factkbgeneralizablefactualityevaluation}
Shangbin Feng, Vidhisha Balachandran, Yuyang Bai, and Yulia Tsvetkov.
\newblock Factkb: Generalizable factuality evaluation using language models enhanced with factual knowledge, 2023{\natexlab{a}}.
\newblock URL \url{https://arxiv.org/abs/2305.08281}.

\bibitem[Feng et~al.(2023{\natexlab{b}})Feng, Park, Liu, and Tsvetkov]{feng2023pretraining}
Shangbin Feng, Chan~Young Park, Yuhan Liu, and Yulia Tsvetkov.
\newblock From pretraining data to language models to downstream tasks: Tracking the trails of political biases leading to unfair nlp models.
\newblock In \emph{Proceedings of the 61st Annual Meeting of the Association for Computational Linguistics (Volume 1: Long Papers)}, pages 11737--11762, 2023{\natexlab{b}}.

\bibitem[Feng et~al.(2024{\natexlab{a}})Feng, Shi, Bai, Balachandran, He, and Tsvetkov]{fengknowledge}
Shangbin Feng, Weijia Shi, Yuyang Bai, Vidhisha Balachandran, Tianxing He, and Yulia Tsvetkov.
\newblock Knowledge card: Filling llms' knowledge gaps with plug-in specialized language models.
\newblock In \emph{The Twelfth International Conference on Learning Representations}, 2024{\natexlab{a}}.

\bibitem[Feng et~al.(2024{\natexlab{b}})Feng, Shi, Wang, Ding, Balachandran, and Tsvetkov]{feng-etal-2024-dont}
Shangbin Feng, Weijia Shi, Yike Wang, Wenxuan Ding, Vidhisha Balachandran, and Yulia Tsvetkov.
\newblock Don{'}t hallucinate, abstain: Identifying {LLM} knowledge gaps via multi-{LLM} collaboration.
\newblock In \emph{Proceedings of the 62nd Annual Meeting of the Association for Computational Linguistics (Volume 1: Long Papers)}, 2024{\natexlab{b}}.

\bibitem[Feng et~al.(2024{\natexlab{c}})Feng, Sorensen, Liu, Fisher, Park, Choi, and Tsvetkov]{feng2024modular}
Shangbin Feng, Taylor Sorensen, Yuhan Liu, Jillian Fisher, Chan~Young Park, Yejin Choi, and Yulia Tsvetkov.
\newblock Modular pluralism: Pluralistic alignment via multi-llm collaboration.
\newblock In \emph{Proceedings of the 2024 Conference on Empirical Methods in Natural Language Processing}, pages 4151--4171, 2024{\natexlab{c}}.

\bibitem[Feng et~al.(2024{\natexlab{d}})Feng, Wang, Wang, Ebrahimi, Palangi, Miculicich, Kulshrestha, Rauschmayr, Choi, Tsvetkov, et~al.]{feng2024model}
Shangbin Feng, Zifeng Wang, Yike Wang, Sayna Ebrahimi, Hamid Palangi, Lesly Miculicich, Achin Kulshrestha, Nathalie Rauschmayr, Yejin Choi, Yulia Tsvetkov, et~al.
\newblock Model swarms: Collaborative search to adapt llm experts via swarm intelligence.
\newblock \emph{arXiv preprint arXiv:2410.11163}, 2024{\natexlab{d}}.

\bibitem[Feng et~al.(2025{\natexlab{a}})Feng, Ding, Liu, Wang, Shi, Wang, Shen, Han, Lang, Lee, Pfister, Choi, and Tsvetkov]{feng2025llmdroolsmultillmcollaboration}
Shangbin Feng, Wenxuan Ding, Alisa Liu, Zifeng Wang, Weijia Shi, Yike Wang, Zejiang Shen, Xiaochuang Han, Hunter Lang, Chen-Yu Lee, Tomas Pfister, Yejin Choi, and Yulia Tsvetkov.
\newblock When one llm drools, multi-llm collaboration rules, 2025{\natexlab{a}}.
\newblock URL \url{https://arxiv.org/abs/2502.04506}.

\bibitem[Feng et~al.(2025{\natexlab{b}})Feng, Shi, Wang, Ding, Ahia, Li, Balachandran, Sitaram, and Tsvetkov]{feng2025teachingllmsabstainlanguages}
Shangbin Feng, Weijia Shi, Yike Wang, Wenxuan Ding, Orevaoghene Ahia, Shuyue~Stella Li, Vidhisha Balachandran, Sunayana Sitaram, and Yulia Tsvetkov.
\newblock Teaching llms to abstain across languages via multilingual feedback, 2025{\natexlab{b}}.
\newblock URL \url{https://arxiv.org/abs/2406.15948}.

\bibitem[Feuer et~al.(2025)Feuer, Goldblum, Datta, Nambiar, Besaleli, Dooley, Cembalest, and Dickerson]{feuer2025styleoutweighssubstancefailure}
Benjamin Feuer, Micah Goldblum, Teresa Datta, Sanjana Nambiar, Raz Besaleli, Samuel Dooley, Max Cembalest, and John~P. Dickerson.
\newblock Style outweighs substance: Failure modes of llm judges in alignment benchmarking, 2025.
\newblock URL \url{https://arxiv.org/abs/2409.15268}.

\bibitem[Fisch et~al.(2024)Fisch, Eisenstein, Zayats, Agarwal, Beirami, Nagpal, Shaw, and Berant]{fisch2024robust}
Adam Fisch, Jacob Eisenstein, Vicky Zayats, Alekh Agarwal, Ahmad Beirami, Chirag Nagpal, Pete Shaw, and Jonathan Berant.
\newblock Robust preference optimization through reward model distillation.
\newblock \emph{arXiv preprint arXiv:2405.19316}, 2024.

\bibitem[Fisher et~al.(2025)Fisher, Appel, Park, Potter, Jiang, Sorensen, Feng, Tsvetkov, Roberts, Pan, Song, and Choi]{fisher2025politicalneutralityaiimpossible}
Jillian Fisher, Ruth~E. Appel, Chan~Young Park, Yujin Potter, Liwei Jiang, Taylor Sorensen, Shangbin Feng, Yulia Tsvetkov, Margaret~E. Roberts, Jennifer Pan, Dawn Song, and Yejin Choi.
\newblock Political neutrality in ai is impossible- but here is how to approximate it, 2025.
\newblock URL \url{https://arxiv.org/abs/2503.05728}.

\bibitem[Gemini~Team et~al.(2023)Gemini~Team, Anil, Borgeaud, Wu, Alayrac, Yu, Soricut, Schalkwyk, Dai, Hauth, et~al.]{team2023gemini}
Gemini Gemini~Team, Rohan Anil, Sebastian Borgeaud, Yonghui Wu, Jean-Baptiste Alayrac, Jiahui Yu, Radu Soricut, Johan Schalkwyk, Andrew~M Dai, Anja Hauth, et~al.
\newblock Gemini: a family of highly capable multimodal models.
\newblock \emph{arXiv preprint arXiv:2312.11805}, 2023.

\bibitem[Guo et~al.(2024)Guo, Zhang, Liu, Liu, Khalman, Llinares, Rame, Mesnard, Zhao, Piot, et~al.]{guo2024direct}
Shangmin Guo, Biao Zhang, Tianlin Liu, Tianqi Liu, Misha Khalman, Felipe Llinares, Alexandre Rame, Thomas Mesnard, Yao Zhao, Bilal Piot, et~al.
\newblock Direct language model alignment from online ai feedback.
\newblock \emph{arXiv preprint arXiv:2402.04792}, 2024.

\bibitem[Hendrycks et~al.(2021)Hendrycks, Burns, Basart, Zou, Mazeika, Song, and Steinhardt]{hendrycksmeasuring}
Dan Hendrycks, Collin Burns, Steven Basart, Andy Zou, Mantas Mazeika, Dawn Song, and Jacob Steinhardt.
\newblock Measuring massive multitask language understanding.
\newblock In \emph{International Conference on Learning Representations}, 2021.

\bibitem[Hendrycks et~al.(2023)Hendrycks, Burns, Basart, Critch, Li, Song, and Steinhardt]{hendrycks2023aligningaisharedhuman}
Dan Hendrycks, Collin Burns, Steven Basart, Andrew Critch, Jerry Li, Dawn Song, and Jacob Steinhardt.
\newblock Aligning ai with shared human values, 2023.
\newblock URL \url{https://arxiv.org/abs/2008.02275}.

\bibitem[Hu et~al.(2022)Hu, Wallis, Allen-Zhu, Li, Wang, Wang, Chen, et~al.]{hulora}
Edward~J Hu, Phillip Wallis, Zeyuan Allen-Zhu, Yuanzhi Li, Shean Wang, Lu~Wang, Weizhu Chen, et~al.
\newblock Lora: Low-rank adaptation of large language models.
\newblock In \emph{International Conference on Learning Representations}, 2022.

\bibitem[Huang et~al.(2022)Huang, Gu, Hou, Wu, Wang, Yu, and Han]{huang2022large}
Jiaxin Huang, Shixiang~Shane Gu, Le~Hou, Yuexin Wu, Xuezhi Wang, Hongkun Yu, and Jiawei Han.
\newblock Large language models can self-improve.
\newblock \emph{arXiv preprint arXiv:2210.11610}, 2022.

\bibitem[Ibarz et~al.(2018)Ibarz, Leike, Pohlen, Irving, Legg, and Amodei]{ibarz2018reward}
Borja Ibarz, Jan Leike, Tobias Pohlen, Geoffrey Irving, Shane Legg, and Dario Amodei.
\newblock Reward learning from human preferences and demonstrations in atari.
\newblock \emph{Advances in neural information processing systems}, 31, 2018.

\bibitem[Ivison et~al.(2023)Ivison, Wang, Pyatkin, Lambert, Peters, Dasigi, Jang, Wadden, Smith, Beltagy, et~al.]{ivison2023camels}
Hamish Ivison, Yizhong Wang, Valentina Pyatkin, Nathan Lambert, Matthew Peters, Pradeep Dasigi, Joel Jang, David Wadden, Noah~A Smith, Iz~Beltagy, et~al.
\newblock Camels in a changing climate: Enhancing lm adaptation with tulu 2.
\newblock \emph{arXiv preprint arXiv:2311.10702}, 2023.

\bibitem[Jang et~al.(2023)Jang, Kim, Lin, Wang, Hessel, Zettlemoyer, Hajishirzi, Choi, and Ammanabrolu]{jang2023personalized}
Joel Jang, Seungone Kim, Bill~Yuchen Lin, Yizhong Wang, Jack Hessel, Luke Zettlemoyer, Hannaneh Hajishirzi, Yejin Choi, and Prithviraj Ammanabrolu.
\newblock Personalized soups: Personalized large language model alignment via post-hoc parameter merging.
\newblock \emph{arXiv preprint arXiv:2310.11564}, 2023.

\bibitem[Ji et~al.(2023)Ji, Lee, Frieske, Yu, Su, Xu, Ishii, Bang, Madotto, and Fung]{ji2023survey}
Ziwei Ji, Nayeon Lee, Rita Frieske, Tiezheng Yu, Dan Su, Yan Xu, Etsuko Ishii, Ye~Jin Bang, Andrea Madotto, and Pascale Fung.
\newblock Survey of hallucination in natural language generation.
\newblock \emph{ACM Computing Surveys}, 55\penalty0 (12):\penalty0 1--38, 2023.

\bibitem[Jiang et~al.(2023)Jiang, Ren, and Lin]{jiang2023llm}
Dongfu Jiang, Xiang Ren, and Bill~Yuchen Lin.
\newblock Llm-blender: Ensembling large language models with pairwise ranking and generative fusion.
\newblock In \emph{Proceedings of the 61st Annual Meeting of the Association for Computational Linguistics (Volume 1: Long Papers)}, pages 14165--14178, 2023.

\bibitem[Jiang et~al.(2024)Jiang, Li, Zhang, Huang, Lin, and Chen]{jiang2024tigerscore}
Dongfu Jiang, Yishan Li, Ge~Zhang, Wenhao Huang, Bill~Yuchen Lin, and Wenhu Chen.
\newblock {TIGERS}core: Towards building explainable metric for all text generation tasks.
\newblock \emph{Transactions on Machine Learning Research (TMLR)}, May 2024.
\newblock ISSN 2835-8856.
\newblock URL \url{https://openreview.net/forum?id=EE1CBKC0SZ}.

\bibitem[Jin et~al.(2021)Jin, Pan, Oufattole, Weng, Fang, and Szolovits]{jin2021disease}
Di~Jin, Eileen Pan, Nassim Oufattole, Wei-Hung Weng, Hanyi Fang, and Peter Szolovits.
\newblock What disease does this patient have? a large-scale open domain question answering dataset from medical exams.
\newblock \emph{Applied Sciences}, 11\penalty0 (14):\penalty0 6421, 2021.

\bibitem[Kasai et~al.(2024)Kasai, Sakaguchi, Le~Bras, Asai, Yu, Radev, Smith, Choi, Inui, et~al.]{kasai2024realtime}
Jungo Kasai, Keisuke Sakaguchi, Ronan Le~Bras, Akari Asai, Xinyan Yu, Dragomir Radev, Noah~A Smith, Yejin Choi, Kentaro Inui, et~al.
\newblock Realtime qa: what's the answer right now?
\newblock \emph{Advances in Neural Information Processing Systems}, 36, 2024.

\bibitem[Kirk et~al.(2023)Kirk, Mediratta, Nalmpantis, Luketina, Hambro, Grefenstette, and Raileanu]{kirk2023understanding}
Robert Kirk, Ishita Mediratta, Christoforos Nalmpantis, Jelena Luketina, Eric Hambro, Edward Grefenstette, and Roberta Raileanu.
\newblock Understanding the effects of rlhf on llm generalisation and diversity.
\newblock \emph{arXiv preprint arXiv:2310.06452}, 2023.

\bibitem[Kiruluta et~al.(2025)Kiruluta, Lemos, and Burity]{kiruluta2025selfsupervisedreinforcementlearningapproach}
Andrew Kiruluta, Andreas Lemos, and Priscilla Burity.
\newblock A self-supervised reinforcement learning approach for fine-tuning large language models using cross-attention signals, 2025.
\newblock URL \url{https://arxiv.org/abs/2502.10482}.

\bibitem[Ko et~al.(2025)Ko, Dingliwal, Ganesh, Sengupta, Bodapati, and Galstyan]{ko2025sera}
Jongwoo Ko, Saket Dingliwal, Bhavana Ganesh, Sailik Sengupta, Sravan~Babu Bodapati, and Aram Galstyan.
\newblock Se{RA}: Self-reviewing and alignment of {LLM}s using implicit reward margins.
\newblock In \emph{The Thirteenth International Conference on Learning Representations}, 2025.
\newblock URL \url{https://openreview.net/forum?id=uIGnuyDSB9}.

\bibitem[K{\"o}pf et~al.(2024)K{\"o}pf, Kilcher, von R{\"u}tte, Anagnostidis, Tam, Stevens, Barhoum, Nguyen, Stanley, Nagyfi, et~al.]{kopf2024openassistant}
Andreas K{\"o}pf, Yannic Kilcher, Dimitri von R{\"u}tte, Sotiris Anagnostidis, Zhi~Rui Tam, Keith Stevens, Abdullah Barhoum, Duc Nguyen, Oliver Stanley, Rich{\'a}rd Nagyfi, et~al.
\newblock Openassistant conversations-democratizing large language model alignment.
\newblock \emph{Advances in Neural Information Processing Systems}, 36, 2024.

\bibitem[Lazaridou et~al.(2021)Lazaridou, Kuncoro, Gribovskaya, Agrawal, Liska, Terzi, Gimenez, de~Masson~d’Autume, Ruder, Yogatama, et~al.]{lazaridou2021pitfalls}
Angeliki Lazaridou, Adhiguna Kuncoro, Elena Gribovskaya, Devang Agrawal, Adam Liska, Tayfun Terzi, Mai Gimenez, Cyprien de~Masson~d’Autume, Sebastian Ruder, Dani Yogatama, et~al.
\newblock Pitfalls of static language modelling.
\newblock \emph{arXiv preprint arXiv:2102.01951}, 2021.

\bibitem[Leibo et~al.(2024)Leibo, Vezhnevets, Diaz, Agapiou, Cunningham, Sunehag, Haas, Koster, Du{\'e}{\~n}ez-Guzm{\'a}n, Isaac, et~al.]{leibo2024theory}
Joel~Z Leibo, Alexander~Sasha Vezhnevets, Manfred Diaz, John~P Agapiou, William~A Cunningham, Peter Sunehag, Julia Haas, Raphael Koster, Edgar~A Du{\'e}{\~n}ez-Guzm{\'a}n, William~S Isaac, et~al.
\newblock A theory of appropriateness with applications to generative artificial intelligence.
\newblock \emph{arXiv preprint arXiv:2412.19010}, 2024.

\bibitem[Leike et~al.(2018)Leike, Krueger, Everitt, Martic, Maini, and Legg]{leike2018scalable}
Jan Leike, David Krueger, Tom Everitt, Miljan Martic, Vishal Maini, and Shane Legg.
\newblock Scalable agent alignment via reward modeling: a research direction.
\newblock \emph{arXiv preprint arXiv:1811.07871}, 2018.

\bibitem[Li et~al.(2024{\natexlab{a}})Li, Ding, Karten, and Jin]{li2024fightladderbenchmarkcompetitivemultiagent}
Wenzhe Li, Zihan Ding, Seth Karten, and Chi Jin.
\newblock Fightladder: A benchmark for competitive multi-agent reinforcement learning, 2024{\natexlab{a}}.
\newblock URL \url{https://arxiv.org/abs/2406.02081}.

\bibitem[Li et~al.(2024{\natexlab{b}})Li, Zheran~Liu, Liang, and Hashimoto]{li2024autobencher}
Xiang~Lisa Li, Evan Zheran~Liu, Percy Liang, and Tatsunori Hashimoto.
\newblock Autobencher: Creating salient, novel, difficult datasets for language models.
\newblock \emph{arXiv e-prints}, pages arXiv--2407, 2024{\natexlab{b}}.

\bibitem[Lian et~al.(2023)Lian, Goodson, Pentland, et~al.]{lian2023openorca}
W~Lian, B~Goodson, E~Pentland, et~al.
\newblock Openorca: An open dataset of gpt augmented flan reasoning traces, 2023.

\bibitem[Liang et~al.(2023)Liang, He, Jiao, Wang, Wang, Wang, Yang, Shi, and Tu]{liang2023encouraging}
Tian Liang, Zhiwei He, Wenxiang Jiao, Xing Wang, Yan Wang, Rui Wang, Yujiu Yang, Shuming Shi, and Zhaopeng Tu.
\newblock Encouraging divergent thinking in large language models through multi-agent debate.
\newblock \emph{arXiv preprint arXiv:2305.19118}, 2023.

\bibitem[Lin et~al.(2022)Lin, Hilton, and Evans]{lin2022truthfulqa}
Stephanie Lin, Jacob Hilton, and Owain Evans.
\newblock Truthfulqa: Measuring how models mimic human falsehoods.
\newblock In \emph{Proceedings of the 60th Annual Meeting of the Association for Computational Linguistics (Volume 1: Long Papers)}, pages 3214--3252, 2022.

\bibitem[Lindner et~al.(2021)Lindner, Turchetta, Tschiatschek, Ciosek, and Krause]{lindner2021information}
David Lindner, Matteo Turchetta, Sebastian Tschiatschek, Kamil Ciosek, and Andreas Krause.
\newblock Information directed reward learning for reinforcement learning.
\newblock \emph{Advances in Neural Information Processing Systems}, 34:\penalty0 3850--3862, 2021.

\bibitem[Madaan et~al.(2024)Madaan, Tandon, Gupta, Hallinan, Gao, Wiegreffe, Alon, Dziri, Prabhumoye, Yang, et~al.]{madaan2024self}
Aman Madaan, Niket Tandon, Prakhar Gupta, Skyler Hallinan, Luyu Gao, Sarah Wiegreffe, Uri Alon, Nouha Dziri, Shrimai Prabhumoye, Yiming Yang, et~al.
\newblock Self-refine: Iterative refinement with self-feedback.
\newblock \emph{Advances in Neural Information Processing Systems}, 36, 2024.

\bibitem[Magee and Galinsky(2008{\natexlab{a}})]{magee20088}
Joe~C Magee and Adam~D Galinsky.
\newblock 8 social hierarchy: The self-reinforcing nature of power and status.
\newblock \emph{The academy of management annals}, 2\penalty0 (1):\penalty0 351--398, 2008{\natexlab{a}}.

\bibitem[Magee and Galinsky(2008{\natexlab{b}})]{magee2008self}
Joseph~C Magee and Adam~D Galinsky.
\newblock The self-reinforcing nature of social hierarchy: Origins and consequences of power and status.
\newblock In \emph{IACM 21st annual conference paper}, 2008{\natexlab{b}}.

\bibitem[Meng et~al.(2024)Meng, Xia, and Chen]{meng2024simpo}
Yu~Meng, Mengzhou Xia, and Danqi Chen.
\newblock Simpo: Simple preference optimization with a reference-free reward.
\newblock \emph{arXiv preprint arXiv:2405.14734}, 2024.

\bibitem[Min et~al.(2023)Min, Krishna, Lyu, Lewis, Yih, Koh, Iyyer, Zettlemoyer, and Hajishirzi]{factscore}
Sewon Min, Kalpesh Krishna, Xinxi Lyu, Mike Lewis, Wen-tau Yih, Pang~Wei Koh, Mohit Iyyer, Luke Zettlemoyer, and Hannaneh Hajishirzi.
\newblock {FActScore}: Fine-grained atomic evaluation of factual precision in long form text generation.
\newblock In \emph{EMNLP}, 2023.
\newblock URL \url{https://arxiv.org/abs/2305.14251}.

\bibitem[Naik et~al.(2024)Naik, Chandrasekaran, Yuksekgonul, Palangi, and Nushi]{naik2024diversity}
Ranjita Naik, Varun Chandrasekaran, Mert Yuksekgonul, Hamid Palangi, and Besmira Nushi.
\newblock {DIVERSITY} {OF} {THOUGHT} {IMPROVES} {REASONING} {ABILITIES} {OF} {LARGE} {LANGUAGE} {MODELS}, 2024.
\newblock URL \url{https://openreview.net/forum?id=FvfhHucpLd}.

\bibitem[Naous et~al.(2023)Naous, Ryan, Ritter, and Xu]{naous2023having}
Tarek Naous, Michael~J Ryan, Alan Ritter, and Wei Xu.
\newblock Having beer after prayer? measuring cultural bias in large language models.
\newblock \emph{arXiv preprint arXiv:2305.14456}, 2023.

\bibitem[Ouyang et~al.(2022)Ouyang, Wu, Jiang, Almeida, Wainwright, Mishkin, Zhang, Agarwal, Slama, Ray, et~al.]{ouyang2022training}
Long Ouyang, Jeffrey Wu, Xu~Jiang, Diogo Almeida, Carroll Wainwright, Pamela Mishkin, Chong Zhang, Sandhini Agarwal, Katarina Slama, Alex Ray, et~al.
\newblock Training language models to follow instructions with human feedback.
\newblock \emph{Advances in neural information processing systems}, 35:\penalty0 27730--27744, 2022.

\bibitem[Pang et~al.(2024)Pang, Tang, Ye, Xiong, Zhang, Wang, and Chen]{matrix_icml2024}
Xianghe Pang, Shuo Tang, Rui Ye, Yuxin Xiong, Bolun Zhang, Yanfeng Wang, and Siheng Chen.
\newblock Self-alignment of large language models via monopolylogue-based social scene simulation.
\newblock In \emph{Proceedings of the 41st International Conference on Machine Learning}, 2024.

\bibitem[Qwen et~al.(2025)Qwen, :, Yang, Yang, Zhang, Hui, Zheng, Yu, Li, Liu, Huang, Wei, Lin, Yang, Tu, Zhang, Yang, Yang, Zhou, Lin, Dang, Lu, Bao, Yang, Yu, Li, Xue, Zhang, Zhu, Men, Lin, Li, Tang, Xia, Ren, Ren, Fan, Su, Zhang, Wan, Liu, Cui, Zhang, and Qiu]{qwen2025qwen25technicalreport}
Qwen, :, An~Yang, Baosong Yang, Beichen Zhang, Binyuan Hui, Bo~Zheng, Bowen Yu, Chengyuan Li, Dayiheng Liu, Fei Huang, Haoran Wei, Huan Lin, Jian Yang, Jianhong Tu, Jianwei Zhang, Jianxin Yang, Jiaxi Yang, Jingren Zhou, Junyang Lin, Kai Dang, Keming Lu, Keqin Bao, Kexin Yang, Le~Yu, Mei Li, Mingfeng Xue, Pei Zhang, Qin Zhu, Rui Men, Runji Lin, Tianhao Li, Tianyi Tang, Tingyu Xia, Xingzhang Ren, Xuancheng Ren, Yang Fan, Yang Su, Yichang Zhang, Yu~Wan, Yuqiong Liu, Zeyu Cui, Zhenru Zhang, and Zihan Qiu.
\newblock Qwen2.5 technical report, 2025.
\newblock URL \url{https://arxiv.org/abs/2412.15115}.

\bibitem[Rafailov et~al.(2023)Rafailov, Sharma, Mitchell, Manning, Ermon, and Finn]{dpo}
Rafael Rafailov, Archit Sharma, Eric Mitchell, Christopher~D. Manning, Stefano Ermon, and Chelsea Finn.
\newblock Direct preference optimization: Your language model is secretly a reward model.
\newblock In \emph{Advances in Neural Information Processing Systems 36: Annual Conference on Neural Information Processing Systems 2023, NeurIPS 2023, New Orleans, LA, USA, December 10 - 16, 2023}, 2023.

\bibitem[Rafailov et~al.(2024)Rafailov, Sharma, Mitchell, Manning, Ermon, and Finn]{rafailov2024direct}
Rafael Rafailov, Archit Sharma, Eric Mitchell, Christopher~D Manning, Stefano Ermon, and Chelsea Finn.
\newblock self-repreference optimization: Your language model is secretly a reward model.
\newblock \emph{Advances in Neural Information Processing Systems}, 36, 2024.

\bibitem[Rao et~al.(2024)Rao, Yerukola, Shah, Reinecke, and Sap]{rao2024normad}
Abhinav Rao, Akhila Yerukola, Vishwa Shah, Katharina Reinecke, and Maarten Sap.
\newblock Normad: A benchmark for measuring the cultural adaptability of large language models.
\newblock \emph{arXiv preprint arXiv:2404.12464}, 2024.

\bibitem[Santurkar et~al.(2023)Santurkar, Durmus, Ladhak, Lee, Liang, and Hashimoto]{santurkar2023whose}
Shibani Santurkar, Esin Durmus, Faisal Ladhak, Cinoo Lee, Percy Liang, and Tatsunori Hashimoto.
\newblock Whose opinions do language models reflect?
\newblock In \emph{International Conference on Machine Learning}, pages 29971--30004. PMLR, 2023.

\bibitem[Sauder et~al.(2012)Sauder, Lynn, and Podolny]{sauder2012status}
Michael Sauder, Freda Lynn, and Joel~M Podolny.
\newblock Status: Insights from organizational sociology.
\newblock \emph{Annual Review of Sociology}, 38\penalty0 (1):\penalty0 267--283, 2012.

\bibitem[Schulman et~al.(2017)Schulman, Wolski, Dhariwal, Radford, and Klimov]{schulman2017proximal}
John Schulman, Filip Wolski, Prafulla Dhariwal, Alec Radford, and Oleg Klimov.
\newblock Proximal policy optimization algorithms.
\newblock \emph{arXiv preprint arXiv:1707.06347}, 2017.

\bibitem[Shen et~al.(2023)Shen, Chen, Song, Jin, Peng, Mi, Khashabi, and Yu]{reward_inconsistency}
Lingfeng Shen, Sihao Chen, Linfeng Song, Lifeng Jin, Baolin Peng, Haitao Mi, Daniel Khashabi, and Dong Yu.
\newblock The trickle-down impact of reward (in-)consistency on {RLHF}.
\newblock \emph{CoRR}, abs/2309.16155, 2023.

\bibitem[Shi et~al.(2024)Shi, Li, Zhang, Ziems, Horesh, de~Paula, Yang, et~al.]{shi2024culturebank}
Weiyan Shi, Ryan Li, Yutong Zhang, Caleb Ziems, Raya Horesh, Rog{\'e}rio~Abreu de~Paula, Diyi Yang, et~al.
\newblock Culturebank: An online community-driven knowledge base towards culturally aware language technologies.
\newblock \emph{arXiv preprint arXiv:2404.15238}, 2024.

\bibitem[Shinn et~al.(2024)Shinn, Cassano, Gopinath, Narasimhan, and Yao]{shinn2024reflexion}
Noah Shinn, Federico Cassano, Ashwin Gopinath, Karthik Narasimhan, and Shunyu Yao.
\newblock Reflexion: Language agents with verbal reinforcement learning.
\newblock \emph{Advances in Neural Information Processing Systems}, 36, 2024.

\bibitem[Slocum et~al.(2025)Slocum, Parker-Sartori, and Hadfield-Menell]{slocum2025diverse}
Stewart Slocum, Asher Parker-Sartori, and Dylan Hadfield-Menell.
\newblock Diverse preference learning for capabilities and alignment.
\newblock In \emph{The Thirteenth International Conference on Learning Representations}, 2025.

\bibitem[Song et~al.(2024)Song, Yu, Li, Yu, Huang, Li, and Wang]{song2024preference}
Feifan Song, Bowen Yu, Minghao Li, Haiyang Yu, Fei Huang, Yongbin Li, and Houfeng Wang.
\newblock Preference ranking optimization for human alignment.
\newblock In \emph{Proceedings of the Thirty-Eighth AAAI Conference on Artificial Intelligence and Thirty-Sixth Conference on Innovative Applications of Artificial Intelligence and Fourteenth Symposium on Educational Advances in Artificial Intelligence}, pages 18990--18998, 2024.

\bibitem[Song et~al.(2023)Song, Khanuja, Liu, Faisal, Ostapenko, Winata, Aji, Cahyawijaya, Tsvetkov, Anastasopoulos, et~al.]{song2023globalbench}
Yueqi Song, Simran Khanuja, Pengfei Liu, Fahim Faisal, Alissa Ostapenko, Genta Winata, Alham Aji, Samuel Cahyawijaya, Yulia Tsvetkov, Antonios Anastasopoulos, et~al.
\newblock Globalbench: A benchmark for global progress in natural language processing.
\newblock In \emph{Proceedings of the 2023 Conference on Empirical Methods in Natural Language Processing}, pages 14157--14171, 2023.

\bibitem[Sorensen et~al.(2024)Sorensen, Moore, Fisher, Gordon, Mireshghallah, Rytting, Ye, Jiang, Lu, Dziri, et~al.]{sorensenposition}
Taylor Sorensen, Jared Moore, Jillian Fisher, Mitchell~L Gordon, Niloofar Mireshghallah, Christopher~Michael Rytting, Andre Ye, Liwei Jiang, Ximing Lu, Nouha Dziri, et~al.
\newblock Position: A roadmap to pluralistic alignment.
\newblock In \emph{Forty-first International Conference on Machine Learning}, 2024.

\bibitem[Subramaniam et~al.(2025)Subramaniam, Du, Tenenbaum, Torralba, Li, and Mordatch]{subramaniam2025multiagent}
Vighnesh Subramaniam, Yilun Du, Joshua~B Tenenbaum, Antonio Torralba, Shuang Li, and Igor Mordatch.
\newblock Multiagent finetuning: Self improvement with diverse reasoning chains.
\newblock \emph{arXiv preprint arXiv:2501.05707}, 2025.

\bibitem[Sun et~al.(2024{\natexlab{a}})Sun, Shen, Cao, Liu, Li, Shen, Gan, Gui, Wang, Yang, Keutzer, and Darrell]{sun-etal-2024-aligning}
Zhiqing Sun, Sheng Shen, Shengcao Cao, Haotian Liu, Chunyuan Li, Yikang Shen, Chuang Gan, Liangyan Gui, Yu-Xiong Wang, Yiming Yang, Kurt Keutzer, and Trevor Darrell.
\newblock Aligning large multimodal models with factually augmented {RLHF}.
\newblock In Lun-Wei Ku, Andre Martins, and Vivek Srikumar, editors, \emph{Findings of the Association for Computational Linguistics: ACL 2024}, pages 13088--13110, Bangkok, Thailand, August 2024{\natexlab{a}}. Association for Computational Linguistics.
\newblock \doi{10.18653/v1/2024.findings-acl.775}.
\newblock URL \url{https://aclanthology.org/2024.findings-acl.775/}.

\bibitem[Sun et~al.(2024{\natexlab{b}})Sun, Yu, Shen, Liu, Yang, Welleck, and Gan]{sun2024easytohard}
Zhiqing Sun, Longhui Yu, Yikang Shen, Weiyang Liu, Yiming Yang, Sean Welleck, and Chuang Gan.
\newblock Easy-to-hard generalization: Scalable alignment beyond human supervision.
\newblock In \emph{The Thirty-eighth Annual Conference on Neural Information Processing Systems}, 2024{\natexlab{b}}.
\newblock URL \url{https://openreview.net/forum?id=qwgfh2fTtN}.

\bibitem[Tao and Li(2025)]{tao2025weakllmsecretlystrong}
Leitian Tao and Yixuan Li.
\newblock Your weak llm is secretly a strong teacher for alignment, 2025.
\newblock URL \url{https://arxiv.org/abs/2409.08813}.

\bibitem[Wan et~al.(2024)Wan, Feng, Tan, Wang, Tsvetkov, and Luo]{wan2024dellgeneratingreactionsexplanations}
Herun Wan, Shangbin Feng, Zhaoxuan Tan, Heng Wang, Yulia Tsvetkov, and Minnan Luo.
\newblock Dell: Generating reactions and explanations for llm-based misinformation detection, 2024.
\newblock URL \url{https://arxiv.org/abs/2402.10426}.

\bibitem[Wang et~al.(2023)Wang, Jiang, Yang, Liu, and Chen]{wang2023reverseklgeneralizingdirect}
Chaoqi Wang, Yibo Jiang, Chenghao Yang, Han Liu, and Yuxin Chen.
\newblock Beyond reverse kl: Generalizing direct preference optimization with diverse divergence constraints, 2023.
\newblock URL \url{https://arxiv.org/abs/2309.16240}.

\bibitem[Wang et~al.(2022{\natexlab{a}})Wang, Wei, Schuurmans, Le, Chi, Narang, Chowdhery, and Zhou]{wang2022self}
Xuezhi Wang, Jason Wei, Dale Schuurmans, Quoc Le, Ed~Chi, Sharan Narang, Aakanksha Chowdhery, and Denny Zhou.
\newblock Self-consistency improves chain of thought reasoning in language models.
\newblock \emph{arXiv preprint arXiv:2203.11171}, 2022{\natexlab{a}}.

\bibitem[Wang et~al.(2022{\natexlab{b}})Wang, Agarwal, Mukherjee, Liu, Gao, Awadallah, and Gao]{wang2022adamixmixtureofadaptationsparameterefficientmodel}
Yaqing Wang, Sahaj Agarwal, Subhabrata Mukherjee, Xiaodong Liu, Jing Gao, Ahmed~Hassan Awadallah, and Jianfeng Gao.
\newblock Adamix: Mixture-of-adaptations for parameter-efficient model tuning, 2022{\natexlab{b}}.

\bibitem[Wang et~al.(2025{\natexlab{a}})Wang, He, Liang, Zhang, Bansal, Wei, Zhang, and Yao]{wang2025creamconsistencyregularizedselfrewarding}
Zhaoyang Wang, Weilei He, Zhiyuan Liang, Xuchao Zhang, Chetan Bansal, Ying Wei, Weitong Zhang, and Huaxiu Yao.
\newblock Cream: Consistency regularized self-rewarding language models, 2025{\natexlab{a}}.
\newblock URL \url{https://arxiv.org/abs/2410.12735}.

\bibitem[Wang et~al.(2025{\natexlab{b}})Wang, Wang, Wang, Zhang, Li, Yang, Yu, Nguyen, Liu, Gottlieb, Lam, Lu, Cho, Wu, Fei-Fei, Wang, Choi, and Li]{wang2025ragenunderstandingselfevolutionllm}
Zihan Wang, Kangrui Wang, Qineng Wang, Pingyue Zhang, Linjie Li, Zhengyuan Yang, Kefan Yu, Minh~Nhat Nguyen, Licheng Liu, Eli Gottlieb, Monica Lam, Yiping Lu, Kyunghyun Cho, Jiajun Wu, Li~Fei-Fei, Lijuan Wang, Yejin Choi, and Manling Li.
\newblock Ragen: Understanding self-evolution in llm agents via multi-turn reinforcement learning, 2025{\natexlab{b}}.
\newblock URL \url{https://arxiv.org/abs/2504.20073}.

\bibitem[Wu et~al.(2024{\natexlab{a}})Wu, Yuan, Golovneva, Xu, Tian, Jiao, Weston, and Sukhbaatar]{wu2024metarewardinglanguagemodelsselfimproving}
Tianhao Wu, Weizhe Yuan, Olga Golovneva, Jing Xu, Yuandong Tian, Jiantao Jiao, Jason Weston, and Sainbayar Sukhbaatar.
\newblock Meta-rewarding language models: Self-improving alignment with llm-as-a-meta-judge, 2024{\natexlab{a}}.
\newblock URL \url{https://arxiv.org/abs/2407.19594}.

\bibitem[Wu et~al.(2024{\natexlab{b}})Wu, Sun, Yuan, Ji, Yang, and Gu]{wu2024selfplaypreferenceoptimizationlanguage}
Yue Wu, Zhiqing Sun, Huizhuo Yuan, Kaixuan Ji, Yiming Yang, and Quanquan Gu.
\newblock Self-play preference optimization for language model alignment, 2024{\natexlab{b}}.
\newblock URL \url{https://arxiv.org/abs/2405.00675}.

\bibitem[Wu et~al.(2023)Wu, Hu, Shi, Dziri, Suhr, Ammanabrolu, Smith, Ostendorf, and Hajishirzi]{wu2023fine}
Zeqiu Wu, Yushi Hu, Weijia Shi, Nouha Dziri, Alane Suhr, Prithviraj Ammanabrolu, Noah~A Smith, Mari Ostendorf, and Hannaneh Hajishirzi.
\newblock Fine-grained human feedback gives better rewards for language model training.
\newblock \emph{Advances in Neural Information Processing Systems}, 36:\penalty0 59008--59033, 2023.

\bibitem[Xiao et~al.(2024)Xiao, Li, Xie, Getzen, Fang, Long, and Su]{xiao2024algorithmicbiasaligninglarge}
Jiancong Xiao, Ziniu Li, Xingyu Xie, Emily Getzen, Cong Fang, Qi~Long, and Weijie~J. Su.
\newblock On the algorithmic bias of aligning large language models with rlhf: Preference collapse and matching regularization, 2024.
\newblock URL \url{https://arxiv.org/abs/2405.16455}.

\bibitem[Xiong et~al.(2023)Xiong, Ding, Cao, Liu, and Qin]{xiong2023examining}
Kai Xiong, Xiao Ding, Yixin Cao, Ting Liu, and Bing Qin.
\newblock Examining inter-consistency of large language models collaboration: An in-depth analysis via debate.
\newblock \emph{arXiv preprint arXiv:2305.11595}, 2023.

\bibitem[Xu et~al.(2023)Xu, Sun, Zheng, Geng, Zhao, Feng, Tao, and Jiang]{xu2023wizardlm}
Can Xu, Qingfeng Sun, Kai Zheng, Xiubo Geng, Pu~Zhao, Jiazhan Feng, Chongyang Tao, and Daxin Jiang.
\newblock Wizardlm: Empowering large language models to follow complex instructions.
\newblock \emph{arXiv preprint arXiv:2304.12244}, 2023.

\bibitem[Xu et~al.(2024{\natexlab{a}})Xu, Wu, Diao, Liu, Wang, Chen, and Gao]{xu2024sayself}
Tianyang Xu, Shujin Wu, Shizhe Diao, Xiaoze Liu, Xingyao Wang, Yangyi Chen, and Jing Gao.
\newblock Sayself: Teaching llms to express confidence with self-reflective rationales.
\newblock \emph{arXiv preprint arXiv:2405.20974}, 2024{\natexlab{a}}.

\bibitem[Xu et~al.(2024{\natexlab{b}})Xu, Zhu, Zhao, Pan, Li, and Wang]{xu2024prideprejudicellmamplifies}
Wenda Xu, Guanglei Zhu, Xuandong Zhao, Liangming Pan, Lei Li, and William~Yang Wang.
\newblock Pride and prejudice: Llm amplifies self-bias in self-refinement, 2024{\natexlab{b}}.
\newblock URL \url{https://arxiv.org/abs/2402.11436}.

\bibitem[Yao et~al.(2024)Yao, Ding, Feng, Wang, and Tsvetkov]{yao2024varyingshadeswrongaligning}
Jihan Yao, Wenxuan Ding, Shangbin Feng, Lucy~Lu Wang, and Yulia Tsvetkov.
\newblock Varying shades of wrong: Aligning llms with wrong answers only, 2024.
\newblock URL \url{https://arxiv.org/abs/2410.11055}.

\bibitem[Ye et~al.(2025)Ye, Li, Li, Ai, Zhou, Shen, Yan, and LIU]{ye2025learning}
Ziyi Ye, Xiangsheng Li, Qiuchi Li, Qingyao Ai, Yujia Zhou, Wei Shen, Dong Yan, and Yiqun LIU.
\newblock Learning {LLM}-as-a-judge for preference alignment.
\newblock In \emph{The Thirteenth International Conference on Learning Representations}, 2025.
\newblock URL \url{https://openreview.net/forum?id=HZVIQE1MsJ}.

\bibitem[Yuan et~al.(2025)Yuan, Pang, Cho, Li, Sukhbaatar, Xu, and Weston]{yuan2025selfrewardinglanguagemodels}
Weizhe Yuan, Richard~Yuanzhe Pang, Kyunghyun Cho, Xian Li, Sainbayar Sukhbaatar, Jing Xu, and Jason Weston.
\newblock Self-rewarding language models, 2025.
\newblock URL \url{https://arxiv.org/abs/2401.10020}.

\bibitem[Zelikman et~al.(2022)Zelikman, Wu, Mu, and Goodman]{zelikman2022star}
Eric Zelikman, Yuhuai Wu, Jesse Mu, and Noah Goodman.
\newblock Star: Bootstrapping reasoning with reasoning.
\newblock \emph{Advances in Neural Information Processing Systems}, 35:\penalty0 15476--15488, 2022.

\bibitem[Zhao et~al.(2024)Zhao, Plaza-del Arco, and Curry]{zhao2024language}
Justin Zhao, Flor~Miriam Plaza-del Arco, and Amanda~Cercas Curry.
\newblock Language model council: Benchmarking foundation models on highly subjective tasks by consensus.
\newblock \emph{arXiv preprint arXiv:2406.08598}, 2024.

\bibitem[Zheng et~al.(2023)Zheng, Chiang, Sheng, Zhuang, Wu, Zhuang, Lin, Li, Li, Xing, et~al.]{zheng2023judging}
Lianmin Zheng, Wei-Lin Chiang, Ying Sheng, Siyuan Zhuang, Zhanghao Wu, Yonghao Zhuang, Zi~Lin, Zhuohan Li, Dacheng Li, Eric Xing, et~al.
\newblock Judging llm-as-a-judge with mt-bench and chatbot arena.
\newblock \emph{Advances in Neural Information Processing Systems}, 36:\penalty0 46595--46623, 2023.

\bibitem[Zhou et~al.(2024{\natexlab{a}})Zhou, Liu, Xu, Iyer, Sun, Mao, Ma, Efrat, Yu, Yu, et~al.]{zhou2024lima}
Chunting Zhou, Pengfei Liu, Puxin Xu, Srinivasan Iyer, Jiao Sun, Yuning Mao, Xuezhe Ma, Avia Efrat, Ping Yu, Lili Yu, et~al.
\newblock Lima: Less is more for alignment.
\newblock \emph{Advances in Neural Information Processing Systems}, 36, 2024{\natexlab{a}}.

\bibitem[Zhou et~al.(2025)Zhou, Guo, Ma, Gui, Zhang, and Huang]{zhou2025selfconsistencyinternalrewardmodels}
Xin Zhou, Yiwen Guo, Ruotian Ma, Tao Gui, Qi~Zhang, and Xuanjing Huang.
\newblock Self-consistency of the internal reward models improves self-rewarding language models, 2025.
\newblock URL \url{https://arxiv.org/abs/2502.08922}.

\bibitem[Zhou et~al.(2024{\natexlab{b}})Zhou, Fan, Cheng, Yang, Chen, Cui, Wang, Li, Zhang, and Yao]{zhou2024calibrated}
Yiyang Zhou, Zhiyuan Fan, Dongjie Cheng, Sihan Yang, Zhaorun Chen, Chenhang Cui, Xiyao Wang, Yun Li, Linjun Zhang, and Huaxiu Yao.
\newblock Calibrated self-rewarding vision language models.
\newblock \emph{arXiv preprint arXiv:2405.14622}, 2024{\natexlab{b}}.

\bibitem[Ziegler et~al.(2019)Ziegler, Stiennon, Wu, Brown, Radford, Amodei, Christiano, and Irving]{ziegler2019fine}
Daniel~M Ziegler, Nisan Stiennon, Jeffrey Wu, Tom~B Brown, Alec Radford, Dario Amodei, Paul Christiano, and Geoffrey Irving.
\newblock Fine-tuning language models from human preferences.
\newblock \emph{arXiv preprint arXiv:1909.08593}, 2019.

\bibitem[Zuo et~al.(2025)Zuo, Zhang, Qu, Sheng, Zhu, Qi, Sun, Cui, Ding, and Zhou]{zuo2025ttrltesttimereinforcementlearning}
Yuxin Zuo, Kaiyan Zhang, Shang Qu, Li~Sheng, Xuekai Zhu, Biqing Qi, Youbang Sun, Ganqu Cui, Ning Ding, and Bowen Zhou.
\newblock Ttrl: Test-time reinforcement learning, 2025.
\newblock URL \url{https://arxiv.org/abs/2504.16084}.

\end{thebibliography}
\bibliographystyle{plainnat}

\clearpage
\section*{NeurIPS Paper Checklist}

\begin{enumerate}

\item {\bf Claims}
    \item[] Question: Do the main claims made in the abstract and introduction accurately reflect the paper's contributions and scope?
    \item[] Answer: \answerYes{} % Replace by \answerYes{}, \answerNo{}, or \answerNA{}.
    \item[] Justification: Our abstract and introduction include our claims that we use collective language models to achieve self-alignment.
    \item[] Guidelines:
    \begin{itemize}
        \item The answer NA means that the abstract and introduction do not include the claims made in the paper.
        \item The abstract and/or introduction should clearly state the claims made, including the contributions made in the paper and important assumptions and limitations. A No or NA answer to this question will not be perceived well by the reviewers. 
        \item The claims made should match theoretical and experimental results, and reflect how much the results can be expected to generalize to other settings. 
        \item It is fine to include aspirational goals as motivation as long as it is clear that these goals are not attained by the paper. 
    \end{itemize}

\item {\bf Limitations}
    \item[] Question: Does the paper discuss the limitations of the work performed by the authors?
    \item[] Answer: \answerYes{} % Replace by \answerYes{}, \answerNo{}, or \answerNA{}.
    \item[] Justification: Yes! Our paper discuss our limitations in Section~\ref{limit}.
    \item[] Guidelines:
    \begin{itemize}
        \item The answer NA means that the paper has no limitation while the answer No means that the paper has limitations, but those are not discussed in the paper. 
        \item The authors are encouraged to create a separate "Limitations" section in their paper.
        \item The paper should point out any strong assumptions and how robust the results are to violations of these assumptions (e.g., independence assumptions, noiseless settings, model well-specification, asymptotic approximations only holding locally). The authors should reflect on how these assumptions might be violated in practice and what the implications would be.
        \item The authors should reflect on the scope of the claims made, e.g., if the approach was only tested on a few datasets or with a few runs. In general, empirical results often depend on implicit assumptions, which should be articulated.
        \item The authors should reflect on the factors that influence the performance of the approach. For example, a facial recognition algorithm may perform poorly when image resolution is low or images are taken in low lighting. Or a speech-to-text system might not be used reliably to provide closed captions for online lectures because it fails to handle technical jargon.
        \item The authors should discuss the computational efficiency of the proposed algorithms and how they scale with dataset size.
        \item If applicable, the authors should discuss possible limitations of their approach to address problems of privacy and fairness.
        \item While the authors might fear that complete honesty about limitations might be used by reviewers as grounds for rejection, a worse outcome might be that reviewers discover limitations that aren't acknowledged in the paper. The authors should use their best judgment and recognize that individual actions in favor of transparency play an important role in developing norms that preserve the integrity of the community. Reviewers will be specifically instructed to not penalize honesty concerning limitations.
    \end{itemize}

\item {\bf Theory assumptions and proofs}
    \item[] Question: For each theoretical result, does the paper provide the full set of assumptions and a complete (and correct) proof?
    \item[] Answer: \answerYes{} % Replace by \answerYes{}, \answerNo{}, or \answerNA{}.
    \item[] Justification: We provide our assumptions and proof in Section~\ref{method}.
    \item[] Guidelines:
    \begin{itemize}
        \item The answer NA means that the paper does not include theoretical results. 
        \item All the theorems, formulas, and proofs in the paper should be numbered and cross-referenced.
        \item All assumptions should be clearly stated or referenced in the statement of any theorems.
        \item The proofs can either appear in the main paper or the supplemental material, but if they appear in the supplemental material, the authors are encouraged to provide a short proof sketch to provide intuition. 
        \item Inversely, any informal proof provided in the core of the paper should be complemented by formal proofs provided in appendix or supplemental material.
        \item Theorems and Lemmas that the proof relies upon should be properly referenced. 
    \end{itemize}

    \item {\bf Experimental result reproducibility}
    \item[] Question: Does the paper fully disclose all the information needed to reproduce the main experimental results of the paper to the extent that it affects the main claims and/or conclusions of the paper (regardless of whether the code and data are provided or not)?
    \item[] Answer: \answerYes{} % Replace by \answerYes{}, \answerNo{}, or \answerNA{}.
    \item[] Justification: We provide all required information in Section~\ref{experiment}.
    \item[] Guidelines:
    \begin{itemize}
        \item The answer NA means that the paper does not include experiments.
        \item If the paper includes experiments, a No answer to this question will not be perceived well by the reviewers: Making the paper reproducible is important, regardless of whether the code and data are provided or not.
        \item If the contribution is a dataset and/or model, the authors should describe the steps taken to make their results reproducible or verifiable. 
        \item Depending on the contribution, reproducibility can be accomplished in various ways. For example, if the contribution is a novel architecture, describing the architecture fully might suffice, or if the contribution is a specific model and empirical evaluation, it may be necessary to either make it possible for others to replicate the model with the same dataset, or provide access to the model. In general. releasing code and data is often one good way to accomplish this, but reproducibility can also be provided via detailed instructions for how to replicate the results, access to a hosted model (e.g., in the case of a large language model), releasing of a model checkpoint, or other means that are appropriate to the research performed.
        \item While NeurIPS does not require releasing code, the conference does require all submissions to provide some reasonable avenue for reproducibility, which may depend on the nature of the contribution. For example
        \begin{enumerate}
            \item If the contribution is primarily a new algorithm, the paper should make it clear how to reproduce that algorithm.
            \item If the contribution is primarily a new model architecture, the paper should describe the architecture clearly and fully.
            \item If the contribution is a new model (e.g., a large language model), then there should either be a way to access this model for reproducing the results or a way to reproduce the model (e.g., with an open-source dataset or instructions for how to construct the dataset).
            \item We recognize that reproducibility may be tricky in some cases, in which case authors are welcome to describe the particular way they provide for reproducibility. In the case of closed-source models, it may be that access to the model is limited in some way (e.g., to registered users), but it should be possible for other researchers to have some path to reproducing or verifying the results.
        \end{enumerate}
    \end{itemize}

\item {\bf Open access to data and code}
    \item[] Question: Does the paper provide open access to the data and code, with sufficient instructions to faithfully reproduce the main experimental results, as described in supplemental material?
    \item[] Answer: \answerYes{} % Replace by \answerYes{}, \answerNo{}, or \answerNA{}.
    \item[] Justification: We will open access to data and code in open-source platform and submit them via OpenReview.
    \item[] Guidelines:
    \begin{itemize}
        \item The answer NA means that paper does not include experiments requiring code.
        \item Please see the NeurIPS code and data submission guidelines (\url{https://nips.cc/public/guides/CodeSubmissionPolicy}) for more details.
        \item While we encourage the release of code and data, we understand that this might not be possible, so “No” is an acceptable answer. Papers cannot be rejected simply for not including code, unless this is central to the contribution (e.g., for a new open-source benchmark).
        \item The instructions should contain the exact command and environment needed to run to reproduce the results. See the NeurIPS code and data submission guidelines (\url{https://nips.cc/public/guides/CodeSubmissionPolicy}) for more details.
        \item The authors should provide instructions on data access and preparation, including how to access the raw data, preprocessed data, intermediate data, and generated data, etc.
        \item The authors should provide scripts to reproduce all experimental results for the new proposed method and baselines. If only a subset of experiments are reproducible, they should state which ones are omitted from the script and why.
        \item At submission time, to preserve anonymity, the authors should release anonymized versions (if applicable).
        \item Providing as much information as possible in supplemental material (appended to the paper) is recommended, but including URLs to data and code is permitted.
    \end{itemize}

\item {\bf Experimental setting/details}
    \item[] Question: Does the paper specify all the training and test details (e.g., data splits, hyperparameters, how they were chosen, type of optimizer, etc.) necessary to understand the results?
    \item[] Answer: \answerYes{} % Replace by \answerYes{}, \answerNo{}, or \answerNA{}.
    \item[] Justification: We provide the information in Section~\ref{experiment}.
    \item[] Guidelines:
    \begin{itemize}
        \item The answer NA means that the paper does not include experiments.
        \item The experimental setting should be presented in the core of the paper to a level of detail that is necessary to appreciate the results and make sense of them.
        \item The full details can be provided either with the code, in appendix, or as supplemental material.
    \end{itemize}

\item {\bf Experiment statistical significance}
    \item[] Question: Does the paper report error bars suitably and correctly defined or other appropriate information about the statistical significance of the experiments?
    \item[] Answer: \answerYes{} % Replace by \answerYes{}, \answerNo{}, or \answerNA{}.
    \item[] Justification: We provide experiment statistical significance in Section~\ref{results}.
    \item[] Guidelines:
    \begin{itemize}
        \item The answer NA means that the paper does not include experiments.
        \item The authors should answer "Yes" if the results are accompanied by error bars, confidence intervals, or statistical significance tests, at least for the experiments that support the main claims of the paper.
        \item The factors of variability that the error bars are capturing should be clearly stated (for example, train/test split, initialization, random drawing of some parameter, or overall run with given experimental conditions).
        \item The method for calculating the error bars should be explained (closed form formula, call to a library function, bootstrap, etc.)
        \item The assumptions made should be given (e.g., Normally distributed errors).
        \item It should be clear whether the error bar is the standard deviation or the standard error of the mean.
        \item It is OK to report 1-sigma error bars, but one should state it. The authors should preferably report a 2-sigma error bar than state that they have a 96\% CI, if the hypothesis of Normality of errors is not verified.
        \item For asymmetric distributions, the authors should be careful not to show in tables or figures symmetric error bars that would yield results that are out of range (e.g. negative error rates).
        \item If error bars are reported in tables or plots, The authors should explain in the text how they were calculated and reference the corresponding figures or tables in the text.
    \end{itemize}

\item {\bf Experiments compute resources}
    \item[] Question: For each experiment, does the paper provide sufficient information on the computer resources (type of compute workers, memory, time of execution) needed to reproduce the experiments?
    \item[] Answer: \answerYes{} % Replace by \answerYes{}, \answerNo{}, or \answerNA{}.
    \item[] Justification: We provide sufficient information on the computer resources in Section~\ref{experiment}.
    \item[] Guidelines:
    \begin{itemize}
        \item The answer NA means that the paper does not include experiments.
        \item The paper should indicate the type of compute workers CPU or GPU, internal cluster, or cloud provider, including relevant memory and storage.
        \item The paper should provide the amount of compute required for each of the individual experimental runs as well as estimate the total compute. 
        \item The paper should disclose whether the full research project required more compute than the experiments reported in the paper (e.g., preliminary or failed experiments that didn't make it into the paper). 
    \end{itemize}
    
\item {\bf Code of ethics}
    \item[] Question: Does the research conducted in the paper conform, in every respect, with the NeurIPS Code of Ethics \url{https://neurips.cc/public/EthicsGuidelines}?
    \item[] Answer: \answerYes{} % Replace by \answerYes{}, \answerNo{}, or \answerNA{}.
    \item[] Justification: All experiments were conducted following ethical guidelines, including privacy protection and responsible AI development.
    \item[] Guidelines:
    \begin{itemize}
        \item The answer NA means that the authors have not reviewed the NeurIPS Code of Ethics.
        \item If the authors answer No, they should explain the special circumstances that require a deviation from the Code of Ethics.
        \item The authors should make sure to preserve anonymity (e.g., if there is a special consideration due to laws or regulations in their jurisdiction).
    \end{itemize}

\item {\bf Broader impacts}
    \item[] Question: Does the paper discuss both potential positive societal impacts and negative societal impacts of the work performed?
    \item[] Answer: \answerYes{} % Replace by \answerYes{}, \answerNo{}, or \answerNA{}.
    \item[] Justification: We discuss the part in our Introduction.
    \item[] Guidelines:
    \begin{itemize}
        \item The answer NA means that there is no societal impact of the work performed.
        \item If the authors answer NA or No, they should explain why their work has no societal impact or why the paper does not address societal impact.
        \item Examples of negative societal impacts include potential malicious or unintended uses (e.g., disinformation, generating fake profiles, surveillance), fairness considerations (e.g., deployment of technologies that could make decisions that unfairly impact specific groups), privacy considerations, and security considerations.
        \item The conference expects that many papers will be foundational research and not tied to particular applications, let alone deployments. However, if there is a direct path to any negative applications, the authors should point it out. For example, it is legitimate to point out that an improvement in the quality of generative models could be used to generate deepfakes for disinformation. On the other hand, it is not needed to point out that a generic algorithm for optimizing neural networks could enable people to train models that generate Deepfakes faster.
        \item The authors should consider possible harms that could arise when the technology is being used as intended and functioning correctly, harms that could arise when the technology is being used as intended but gives incorrect results, and harms following from (intentional or unintentional) misuse of the technology.
        \item If there are negative societal impacts, the authors could also discuss possible mitigation strategies (e.g., gated release of models, providing defenses in addition to attacks, mechanisms for monitoring misuse, mechanisms to monitor how a system learns from feedback over time, improving the efficiency and accessibility of ML).
    \end{itemize}
    
\item {\bf Safeguards}
    \item[] Question: Does the paper describe safeguards that have been put in place for responsible release of data or models that have a high risk for misuse (e.g., pretrained language models, image generators, or scraped datasets)?
    \item[] Answer: \answerNA{} % Replace by \answerYes{}, \answerNo{}, or \answerNA{}.
    \item[] Justification: We don't use risky data or models.
    \item[] Guidelines:
    \begin{itemize}
        \item The answer NA means that the paper poses no such risks.
        \item Released models that have a high risk for misuse or dual-use should be released with necessary safeguards to allow for controlled use of the model, for example by requiring that users adhere to usage guidelines or restrictions to access the model or implementing safety filters. 
        \item Datasets that have been scraped from the Internet could pose safety risks. The authors should describe how they avoided releasing unsafe images.
        \item We recognize that providing effective safeguards is challenging, and many papers do not require this, but we encourage authors to take this into account and make a best faith effort.
    \end{itemize}

\item {\bf Licenses for existing assets}
    \item[] Question: Are the creators or original owners of assets (e.g., code, data, models), used in the paper, properly credited and are the license and terms of use explicitly mentioned and properly respected?
    \item[] Answer: \answerYes{} % Replace by \answerYes{}, \answerNo{}, or \answerNA{}.
    \item[] Justification: We have cited all the relevant reports or papers in our paper.
    \item[] Guidelines:
    \begin{itemize}
        \item The answer NA means that the paper does not use existing assets.
        \item The authors should cite the original paper that produced the code package or dataset.
        \item The authors should state which version of the asset is used and, if possible, include a URL.
        \item The name of the license (e.g., CC-BY 4.0) should be included for each asset.
        \item For scraped data from a particular source (e.g., website), the copyright and terms of service of that source should be provided.
        \item If assets are released, the license, copyright information, and terms of use in the package should be provided. For popular datasets, \url{paperswithcode.com/datasets} has curated licenses for some datasets. Their licensing guide can help determine the license of a dataset.
        \item For existing datasets that are re-packaged, both the original license and the license of the derived asset (if it has changed) should be provided.
        \item If this information is not available online, the authors are encouraged to reach out to the asset's creators.
    \end{itemize}

\item {\bf New assets}
    \item[] Question: Are new assets introduced in the paper well documented and is the documentation provided alongside the assets?
    \item[] Answer: \answerYes{} % Replace by \answerYes{}, \answerNo{}, or \answerNA{}.
    \item[] Justification: We introduce a new code repositories and we will submit it in our supplementary materials.
    \item[] Guidelines:
    \begin{itemize}
        \item The answer NA means that the paper does not release new assets.
        \item Researchers should communicate the details of the dataset/code/model as part of their submissions via structured templates. This includes details about training, license, limitations, etc. 
        \item The paper should discuss whether and how consent was obtained from people whose asset is used.
        \item At submission time, remember to anonymize your assets (if applicable). You can either create an anonymized URL or include an anonymized zip file.
    \end{itemize}

\item {\bf Crowdsourcing and research with human subjects}
    \item[] Question: For crowdsourcing experiments and research with human subjects, does the paper include the full text of instructions given to participants and screenshots, if applicable, as well as details about compensation (if any)? 
    \item[] Answer: \answerNA{} % Replace by \answerYes{}, \answerNo{}, or \answerNA{}.
    \item[] Justification: We don't involve human in our project.
    \item[] Guidelines:
    \begin{itemize}
        \item The answer NA means that the paper does not involve crowdsourcing nor research with human subjects.
        \item Including this information in the supplemental material is fine, but if the main contribution of the paper involves human subjects, then as much detail as possible should be included in the main paper. 
        \item According to the NeurIPS Code of Ethics, workers involved in data collection, curation, or other labor should be paid at least the minimum wage in the country of the data collector. 
    \end{itemize}

\item {\bf Institutional review board (IRB) approvals or equivalent for research with human subjects}
    \item[] Question: Does the paper describe potential risks incurred by study participants, whether such risks were disclosed to the subjects, and whether Institutional Review Board (IRB) approvals (or an equivalent approval/review based on the requirements of your country or institution) were obtained?
    \item[] Answer: \answerNA{} % Replace by \answerYes{}, \answerNo{}, or \answerNA{}.
    \item[] Justification: We do not involve crowdsourcing nor research with human subjects.
    \item[] Guidelines:
    \begin{itemize}
        \item The answer NA means that the paper does not involve crowdsourcing nor research with human subjects.
        \item Depending on the country in which research is conducted, IRB approval (or equivalent) may be required for any human subjects research. If you obtained IRB approval, you should clearly state this in the paper. 
        \item We recognize that the procedures for this may vary significantly between institutions and locations, and we expect authors to adhere to the NeurIPS Code of Ethics and the guidelines for their institution. 
        \item For initial submissions, do not include any information that would break anonymity (if applicable), such as the institution conducting the review.
    \end{itemize}

\item {\bf Declaration of LLM usage}
    \item[] Question: Does the paper describe the usage of LLMs if it is an important, original, or non-standard component of the core methods in this research? Note that if the LLM is used only for writing, editing, or formatting purposes and does not impact the core methodology, scientific rigorousness, or originality of the research, declaration is not required.
    %this research? 
    \item[] Answer: \answerNA{} % Replace by \answerYes{}, \answerNo{}, or \answerNA{}.
    \item[] Justification: LLMs were not used in the writing of this paper.
    \item[] Guidelines:
    \begin{itemize}
        \item The answer NA means that the core method development in this research does not involve LLMs as any important, original, or non-standard components.
        \item Please refer to our LLM policy (\url{https://neurips.cc/Conferences/2025/LLM}) for what should or should not be described.
    \end{itemize}
\end{enumerate}

\newpage
\appendix

\section{Limitations}\label{limit}
\ourmethod{} selects models for diversity probing experiments based on their initial performance ranking on the validation set. While the optimal combination for maximizing diversity remains unclear, it represents a promising area for future exploration.

Model collaboration in \ourmethod{} requires greater computational resources compared to single models, yet it enables substantial weak-to-strong generalization through synchronized learning, effectively bridging performance gaps across tasks. By selecting the best-performing checkpoint from a pool of 80 based on validation performance, we demonstrate that even initially weaker models can evolve into top performers, underscoring the transformative impact of collective alignment.

Despite \ourmethod{}'s strong performance in reasoning and instruction-following tasks, model diversity diminishes as iterations progress, indicating that improvements may not be infinite. Preserving diversity is crucial for sustained performance gains.

The scaling law persists in \ourmethod{}, showing that larger datasets enable models to better capture latent patterns, underscoring the importance of scale for improved learning and generalization.
\section{Ethics Statement}
The implementation of \ourmethod{} raises several ethical considerations, particularly regarding bias amplification and fairness in collaborative learning. If one of the models in the pool is biased—whether in terms of gender, race, or other sensitive attributes—there is a risk that this bias could be propagated or even amplified through the collective alignment process. This is especially concerning if models with inherent biases gain higher reputation scores due to overfitting to specific patterns or datasets, leading to skewed decision-making.

\section{Technical Appendices}
\subsection{Reweighting Mechanism}\label{reweight}
 To maintain model diversity while preventing `unreliable'' models from affecting the overall judgment so that keeping the preference labels consistent~\citep{wang2025creamconsistencyregularizedselfrewarding,zhou2025selfconsistencyinternalrewardmodels}, we introduce an important attribute: judging weight $w_k$ for $M_k^t$, $M_k^t \notin \{M_i^t, M_{i'}^t\}$ and its corresponding reweighting mechanism. All models are initialized with \(\omega_k = 1\). Starting from iteration \(t = 2\), as training progresses, exactly one additional low-ranked model is down-weighted at each iteration. Specifically, at iteration \(t\), the judge with the \((t - 1)\)-th lowest reputation is unlocked and assigned:
    \[
    \omega_k = \gamma \times (t - 2),
    \]
    where \(\gamma = 0.1\) is a scaling factor. Crucially, once a model is assigned a reduced weight, this weight remains fixed for the remainder of training, regardless of changes in its rank. This progressive and irreversible reweighting penalizes persistent underperformance early while steadily expanding judging participation as alignment stabilizes.

    Thus, the updated reputation $R_k$ for each judge model $M_k^t$:
    \begin{equation*}
       R_k \gets \omega_k\times R_k 
    \end{equation*} 
\subsection{Experiment Dataset Statistics}
The statistics for the datasets used in our main experiments are summarized in Table~\ref{tab:dataset_stats}. The table presents the number of training, validation, and test examples for each of the 12 distinct tasks. Each dataset is split into three parts: Train, Validation, and Test, where the validation set is constructed by splitting the original test set evenly, ensuring balanced evaluation during development and final testing.

For instance, the Alpaca dataset comprises 52,002 training examples, along with 402 validation examples and 403 test examples. Similarly, the MedQA dataset contains 10,178 training samples, with 636 for validation and 637 for testing.

The Normad dataset is divided evenly across three sub-domains: Country, Value, and Rule-of-Thumb (RoT), each containing 2,370 samples in the training set and 132 samples in both validation and test sets. This balanced distribution allows for granular analysis of performance across cultural dimensions.

For the mathematical benchmarks, the MATH dataset is split into three difficulty levels—Easy, Medium, and Hard. The training set includes 1,912, 1,592, and 2,994 samples for each level, respectively. In the validation and test splits, the three levels are proportionally represented, with 666, 566, and 1,269 in the validation set, and 665, 565, and 1,269 in the test set.

Finally, the GSM8K and KC datasets contain 7,473 and 1,200 training samples, respectively. Their validation and test sets are split as 660/659 for GSM8K and 200/200 for KC, respectively.

This structured distribution ensures that each dataset is adequately represented during both development and evaluation phases, providing a fair and consistent basis for assessing model generalization and task-specific performance.
\begin{table}[t]
\centering
\setlength{\tabcolsep}{3pt}
\renewcommand{\arraystretch}{1.1}
\resizebox{1\textwidth}{!}{%
\begin{tabular}{l c c c ccc ccc c c}
\toprule[1.5pt]
\multirow{2}{*}{\textbf{Split}} & 
\multirow{2}{*}{\textbf{Alpaca}} & \multirow{2}{*}{\textbf{MedQA}} & \multirow{2}{*}{\textbf{COM$\textsuperscript{2}$}} & 
\multicolumn{3}{c}{\textbf{Normad}} & 
\multicolumn{3}{c}{\textbf{MATH}} & 
\multirow{2}{*}{\textbf{GSM8K}} & \multirow{2}{*}{\textbf{KC}} \\

\cmidrule(lr){5-7} \cmidrule(lr){8-10}
& & & & \textbf{Country} & \textbf{Value} & \textbf{RoT} & 
\textbf{Easy} & \textbf{Medium} & \textbf{Hard} & & \\
\midrule[0.75pt]

\textsc{Train} & 52,002 & 10,178 & 7,200 & 2,370 & 2,370 & 2,370 & 1,912 & 1,592 & 2,994 & 7,473 & 1,200 \\
\textsc{Val} & 402 & 636 & 400 & 132 & 132 & 132 & 666 & 566 & 1,269 & 660 & 200 \\
\textsc{Test} & 403 & 637 & 400 & 132 & 132 & 132 & 665 & 565 & 1,269 & 659 & 200 \\
\bottomrule[1.5pt]
\end{tabular}
}
\vspace{10pt}
\caption{Dataset statistics for 11 tasks in our main experiments. Validation set is created as half of the original test set, with the remaining portion reserved for final evaluation.}
\vspace*{-10pt}
\label{tab:dataset_stats}
\end{table}

\subsection{SFT Details}\label{sft-data}
We utilize Tulu-v2 \citep{ivison2023camels}, an open collection of instruction-tuning data, focusing specifically on the following subsets: Flan \citep{chung2024scaling}, CoT, Open Assistant 1 \citep{kopf2024openassistant}, ShareGPT\footnote{\url{https://sharegpt.com/}}, Code Alpaca \citep{codealpaca}, LIMA \citep{zhou2024lima}, WizardLM Evol-Instruct V2 \citep{xu2023wizardlm}, Open-Orca \citep{lian2023openorca}, and Science Literature \citep{ivison2023camels}. We replace the GPT4 Alpaca subset with Gemini Alpaca, which consists of distilled generations from \emph{gemini-1.5-pro-001}, and remove the hard-coded subset. Fine-tuning is performed with LoRA \citep{hulora}, employing a learning rate of 2e-4, cosine learning rate scheduling, an effective batch size of 32, a warm-up ratio of 0.1, and 5 default training epochs. For the larger ShareGPT subset, we reduce training to a single epoch to optimize efficiency.

\subsection{Baseline Experiment Settings}\label{baseline-setting}
Inference is conducted with the same inference time and training configurations—including optimization hyperparameters, number of training iterations, and number of sampling prompts—are kept identical across all methods, including \ourmethod{}. This controlled setup isolates the impact of the alignment algorithm, removing potential confounding factors such as model size or training regimen.
\subsection{Experiment Details}
In the Combat and Judge phases of \ourmethod{}, we employ a standardized chat template to generate both responses and judgments. During the DPO training stage, models are trained with the chat template to ensure consistency and alignment. For evaluation, task-specific adjustments are made: Alpaca, COM$^2$, MATH, and GSM8K and MedQA are assessed without the chat template due to the nature of their task structures, while Normad, KCross and Truthful is evaluated with the chat template. Notably, the evaluation methodology remains consistent across the initial model and all baselines.
\paravs
\paragraph{Evaluation Metrics in MedQA}\label{metric-medqa}
In the MedQA evaluation, we employ a multi-faceted assessment strategy with to thoroughly measure the quality of model-generated medical responses. The evaluation is structured around three core metrics: 
\begin{enumerate}
    \item \textbf{Clinical Informativeness:} This metric gauges the depth and richness of medical information provided. A highly informative response should capture critical clinical knowledge, present precise medical terminology, and effectively convey important concepts relevant to the question.

    \item \textbf{Logical Coherence:} This metric assesses the internal consistency and structured reasoning of the response. A logically coherent answer is expected to be well-organized, maintain a clear flow of ideas, and demonstrate sound clinical reasoning without contradictions.

    \item \textbf{Comprehensiveness:} This metric measures the completeness of the response, including coverage of differential diagnoses, treatment options, and underlying mechanisms. It ensures that all aspects of the query are adequately addressed with appropriate detail and medical context.
\end{enumerate}
To ensure a consistent and objective evaluation process, we apply LLM-as-a-judge scoring mechanisms for all three metrics. Below are the specific evaluation prompts used for each criterion.
\begin{center}
\begin{tcolorbox}[colback=blue!5!white, colframe=blue!75!black, width=1.0\textwidth, title=Clinical Informativeness Prompt]
Rate the \textbf{clinical informativeness} of the following medical response on a scale of \textbf{0 to 10} (10 being most informative). Consider the medical information intensity.

\textbf{Question:} \{prompt\}  

\textbf{Response:} \{response\}  

Please output in the format: \texttt{\{'score': score\}}
\end{tcolorbox}
\end{center}

\begin{center} 
\begin{tcolorbox}[colback=blue!5!white, colframe=blue!75!black, width=1.0\textwidth, title=Logical Coherence Prompt]
Rate the \textbf{logical coherence} of the following medical response on a scale of \textbf{0 to 10} (10 being most logical). 

\textbf{When evaluating, consider the following aspects:}
\begin{itemize}
    \item \textbf{Structure:} Is the response well-organized and easy to follow?  
    \item \textbf{Flow:} Are the ideas presented in a logical sequence, with clear connections between points?  
    \item \textbf{Reasoning:} Is the reasoning sound and aligned with medical principles?  
\end{itemize}

\textbf{Question:} \{prompt\}  

\textbf{Response:} \{response\}  

Please output in the format: \texttt{\{'score': score\}}
\end{tcolorbox}
\end{center}
\begin{center}
\begin{tcolorbox}[colback=blue!5!white, colframe=blue!75!black, width=1.0\textwidth, title=Comprehensiveness Prompt]
Rate the \textbf{comprehensiveness} of the following medical response on a scale of \textbf{0 to 10} (10 being the most comprehensive).  

\textbf{When evaluating, consider the following aspects:}
\begin{itemize}
    \item \textbf{Completeness:} Does the response address all relevant aspects of the medical query?  
    \item \textbf{Detail:} Are the explanations sufficiently detailed to convey a clear understanding of the concepts involved?  
    \item \textbf{Coverage:} Does the response encompass the full scope of the question, including differential diagnoses, treatment options, and underlying mechanisms where appropriate?  
\end{itemize}

\textbf{Question:} \{prompt\}  

\textbf{Response:} \{response\}  

Please output in the format: \texttt{\{'score': score\}}
\end{tcolorbox}
\end{center}
\clearpage
\paragraph{Results on MedQA}\label{medqa_results}
\begin{wraptable}{r}{0.5\textwidth}
\scriptsize
\centering
\setlength{\tabcolsep}{3pt}
\renewcommand{\arraystretch}{1.1}
\vspace*{-10pt}
\resizebox{0.5\textwidth}{!}{
\begin{tabular}{lcccc}
\toprule[1pt]
\textbf{Method} & \textbf{Comp.} & \textbf{Info.} & \textbf{Logi.} \\
\midrule[0.75pt]
\textsc{Best Init Model} & 6.69 & \underline{7.43} & 8.05 \\
\textsc{Self-reward} & 4.76 & 6.55 & 7.06 \\
\textsc{Meta-reward} & 6.70 & 7.43 & \underline{8.06} \\
\textsc{Spin} & \textbf{6.75} & 7.36 & 8.02 \\
\textsc{Sppo} & 6.70 & 7.30 & 7.97 \\
\textsc{Sparta} & \underline{6.73} & \textbf{7.57} & \textbf{8.13} \\
\bottomrule[1pt]
\end{tabular}
}
\vspace*{-5pt}
\caption{
Evaluation of MedQA Responses with Multi-Faceted Metrics.
}
\label{tab:ablation-appendix}
\vspace*{-10pt}
\end{wraptable}
Table~\ref{tab:ablation-appendix} presents the evaluation results on the MedQA benchmark, measured across three key metrics other than pass@1 accuracy: {Comprehensiveness (Comp.), Informativeness (Info.), and Logical Consistency (Logi.). These metrics provide a holistic assessment of the model's capacity to generate accurate, complete, and well-structured medical responses.

\ourmethod{} achieves the highest scores in both Informativeness and Logical Consistency, outperforming all baselines. This result indicates that \ourmethod{} not only produces responses rich in medical content but also maintains clear logical flow, which is crucial for clinical reasoning in realistic scenario. 

For Comprehensiveness, \textsc{Spin} secures the highest score of 6.75, with \ourmethod{} closely following at 6.73. This highlights \ourmethod{}'s ability to cover critical aspects of medical queries with substantial detail, ensuring that responses are both thorough and contextually rich.

Overall, \ourmethod{} excels in generating informative and logically sound medical responses, setting a new benchmark for clinical AI systems in terms of depth, clarity, and reasoning. These strengths position \ourmethod{} as a highly competitive model for real-world medical QA applications.
\begin{table}[t]
\centering
\setlength{\tabcolsep}{3pt}
\renewcommand{\arraystretch}{1.1}
\resizebox{\textwidth}{!}{%
\begin{tabular}{l c c ccc}
\toprule[1.5pt]
\multirow{2}{*}{\textbf{Stage}} & 
\multirow{2}{*}{\textbf{COM$\textsuperscript{2}$}} & 
\multirow{2}{*}{\textbf{GSM8K}} & 
\multicolumn{3}{c}{\textbf{Normad}} \\

\cmidrule(lr){4-6}
& & & \textbf{Value} & \textbf{RoT} & \\ 
\midrule[0.75pt]

\textsc{Init Best Model} & \textcolor{red}{code\_alpaca} & \textcolor{red}{gemini\_alpaca} & \textcolor{red}{wizardlm} & \textcolor{red}{wizardlm} \\
\textsc{Last Best Model} & \textcolor{blue}{flan\_v2} & \textcolor{blue}{oasst1} & \textcolor{blue}{oasst1} & \textcolor{blue}{oasst1} \\
\bottomrule[1.5pt]
\end{tabular}
}
\vspace{10pt}
\caption{
Comparison of the best-performing models between the initial and final iterations. Only the tasks with changes are shown. Red indicates the initial best model and blue indicates the final best model.
}
\vspace*{-10pt}
\label{tab:best-models}
\end{table}

\subsection{Details in Generalization}

We evaluate \ourmethod{}'s generalization capability using the MATH benchmark~\citep{hendrycksmeasuring}, which assesses LLMs on their reasoning ability across varying levels of mathematical complexity. The MATH benchmark is organized into three difficulty levels: (1) Easy, (2) Medium, and (3) Hard, representing progressively challenging problem sets aimed at testing the model's understanding and problem-solving depth~\citep{sun2024easytohard}. We fine-tune a pool of models on each subset individually and measure their performance on the two held-out subsets, effectively evaluating cross-difficulty generalization.

The experiment can be divided into two phases: the training phase and the testing phase. In the training phase, models are trained on a specific subset of mathematical problems in the MATH benchmark using \ourmethod{}; in the testing phase, these models are evaluated for their performance on the other two subsets of different difficulty levels.

\ourmethod{} ensures that the model adjusts its problem-solving strategies by leveraging a more comprehensive understanding of both fundamental mathematical concepts and advanced reasoning principles. This alignment process enables the model to generalize effectively to unseen problem sets, improving its accuracy and adaptability across varying levels of complexity.

The results indicate that after applying Sparta Alignment, the model's performance improves across basic concepts, intermediate reasoning, and advanced problem-solving contexts, leading to more accurate and contextually consistent solutions.

\subsection{Details in Scaling Law of Model Pool}\label{scaling}
To ensure fair comparisons and preserve meaningful quality gaps for effective alignment training, we first evaluate all candidate models on validation set of one specific task. Based on their test performance rankings, we construct model pools with the largest internal performance spread—specifically selecting the top 3, 5, and 7 models with maximally divergent test performance. For example, the 3-model setting includes models ranked 1st, 5th, and 10th, providing sufficient variance to support informative comparative feedback. We perform this analysis across three representative benchmarks—Alpaca, COM$^2$ and MedQA—to give a conclusion across both general instruction-following and domain-specific tasks.

To enable a unified comparison across tasks with different scoring scales, we normalize the \textsc{Alpaca} and \textsc{COM}$^2$ scores by dividing them by 10.

\subsection{Further Analysis of Generation Diversity}
\paragraph{Details of Diversity between the chosen and rejected one}\label{preference_diversity}
\begin{wraptable}{r}{0.5\textwidth}
\scriptsize
\centering
\setlength{\tabcolsep}{3pt}
\renewcommand{\arraystretch}{1.1}
\vspace*{-10pt}
\resizebox{0.5\textwidth}{!}{
\begin{tabular}{lcccccc}
\toprule[1pt]
\textbf{Method} & \textbf{Char Ent.} & \textbf{Word Ent.} & \textbf{2-Gram} & \textbf{3-Gram} & \textbf{POS.} & \textbf{Semantic Sim.} \\
\midrule[0.75pt]
\textsc{Self-reward} & 0.029 & \textbf{0.220} & \textbf{0.277} & \textbf{0.327} & \textbf{0.100} & \textcolor{blue}{0.925} \\
\textsc{Sparta}    & \textbf{0.066} & 0.175 & 0.186 & 0.197 & 0.079 & \textcolor{red}{0.710} \\
\bottomrule[1pt]
\end{tabular}
}
\vspace*{-5pt}
\caption{
Diversity and similarity metrics comparison between \textsc{Self-reward} and \textsc{Sparta} on entropy-based and semantic similarity measures. \textsc{Sparta} achieves higher character-level entropy and lower semantic similarity. 
}
\label{tab:entropy-metrics}
\vspace*{-10pt}
\end{wraptable}
We evaluate the diversity between the chosen and the rejected ones for \textsc{Self-reward} and \textsc{Sparta} across multiple entropy-based metrics and semantic similarity measures. Specifically, we measure Char Entropy, Word Entropy, 2-Gram Entropy, 3-Gram Entropy, POS Entropy, and Semantic Similarity.

Table~\ref{tab:entropy-metrics} reveals some conclusions. Firstly, the preference pairs in \textsc{Sparta} exhibit substantially lower semantic similarity compared to those in \textsc{Self-reward}. This indicates larger representational gaps, highlighting opportunities for preference optimization to better approximate the latent target data distribution. 

Secondly, the differences in Character Entropy are more pronounced in \textsc{Sparta}, suggesting that variations between the chosen and rejected samples are partially influenced by disparities in length. Among all entropy-based metrics, Character Entropy is the most sensitive to text length, which amplifies its variance. Moreover, \textsc{Sparta} consistently demonstrates lower values across other entropy metrics compared to \textsc{Self-Reward}, indicating reduced diversity in its generated text. And Further case studies validate this observation, revealing that even with identical extracted answers, the shorter responses are consistently marked as negative. This bias implicitly encourages the model to generate more detailed and comprehensive answers during preference optimization.

\paragraph{Details of Evaluation Metrics}\label{evaluation-metric}

We employ Pairwise Diversity Calculation to quantify cross-response diversity for each prompt. Specifically, generated responses are compared in a pairwise manner across three primary metrics: Edit Distance, BLEU Distance, and Embedding Distance. For each metric, the average pairwise distance is computed to reflect the diversity of responses associated with the prompt. To further enhance the evaluation, we incorporate an LLM-based Judgment Mechanism leveraging the \textsc{Gemini-1.5-Flash} model. This model acts as a judge, providing an independent assessment of semantic, stylistic, and strategic differences across generated responses.
\begin{itemize}[leftmargin=*]
    \item \textbf{Edit Distance (Edit):} This metric measures the surface-level lexical diversity by calculating the normalized character-level edit distance between pairs of generated responses. Specifically, for two responses $r_1$ and $r_2$, the edit distance is defined as the minimum number of character-level insertions, deletions, or substitutions required to transform $r_1$ into $r_2$. The result is normalized by the maximum length of the two responses to ensure comparability across different response lengths. Formally, the edit distance $d_{\text{edit}}$ is given by:
    \[
    d_{\text{edit}}(r_1, r_2) = \frac{\text{EditDistance}(r_1, r_2)}{\max(|r_1|, |r_2|)}
    \]
    Higher values indicate greater surface-level variations, reflecting more diverse lexical choices among the responses.

    \item \textbf{BLEU Distance (BLEU):} This metric assesses the structural or phrasal diversity of generated responses by capturing $n$-gram overlap differences. It is computed as $1 - \text{BLEU}$, where BLEU is the traditional bilingual evaluation understudy score. For two responses, the BLEU score is calculated with equal weight for unigram, bigram, trigram, and four-gram matches. The smoothing function is applied to avoid zero scores for sentences with no overlap. The BLEU distance is formulated as:
    \[
    d_{\text{BLEU}}(r_1, r_2) = 1 - \text{BLEU}(r_1, r_2)
    \]
    A higher BLEU distance suggests more structural or syntactic variations between the responses, enhancing phrasal diversity.

    \item \textbf{Embedding Distance (Embed):} This metric quantifies the semantic distinctness between responses by leveraging a pre-trained Sentence Transformer model (BAAI/bge-large-en-v1.5) to compute embeddings. For each pair of responses, the cosine similarity is measured between their embeddings, and the embedding distance is derived as:
    \[
    d_{\text{embed}}(r_1, r_2) = 1 - \cos(\mathbf{e}_1, \mathbf{e}_2)
    \]
    where $\mathbf{e}_1$ and $\mathbf{e}_2$ are the embedding vectors for $r_1$ and $r_2$, respectively. This value reflects the semantic dissimilarity, with higher scores indicating more substantial differences in meaning.

    \item \textbf{LLM Diversity Score (LLM):} This score is calculated by querying the \textsc{Gemini-1.5-Flash} model with a series of generated responses and evaluating the diversity using a proprietary evaluation mechanism. The model scores each prompt based on semantic, stylistic, and strategic differences, providing an aggregate diversity score on a scale from 1 to 10. This metric not only captures linguistic diversity but also considers deeper contextual shifts and strategic variations in the generated responses.
\end{itemize}
\clearpage
\paragraph{Tracking the Changes of Generation Diversity}
\begin{wrapfigure}{r}{0.5\textwidth}
     \vspace{-5pt}
     \centering
     \includegraphics[width=\linewidth]{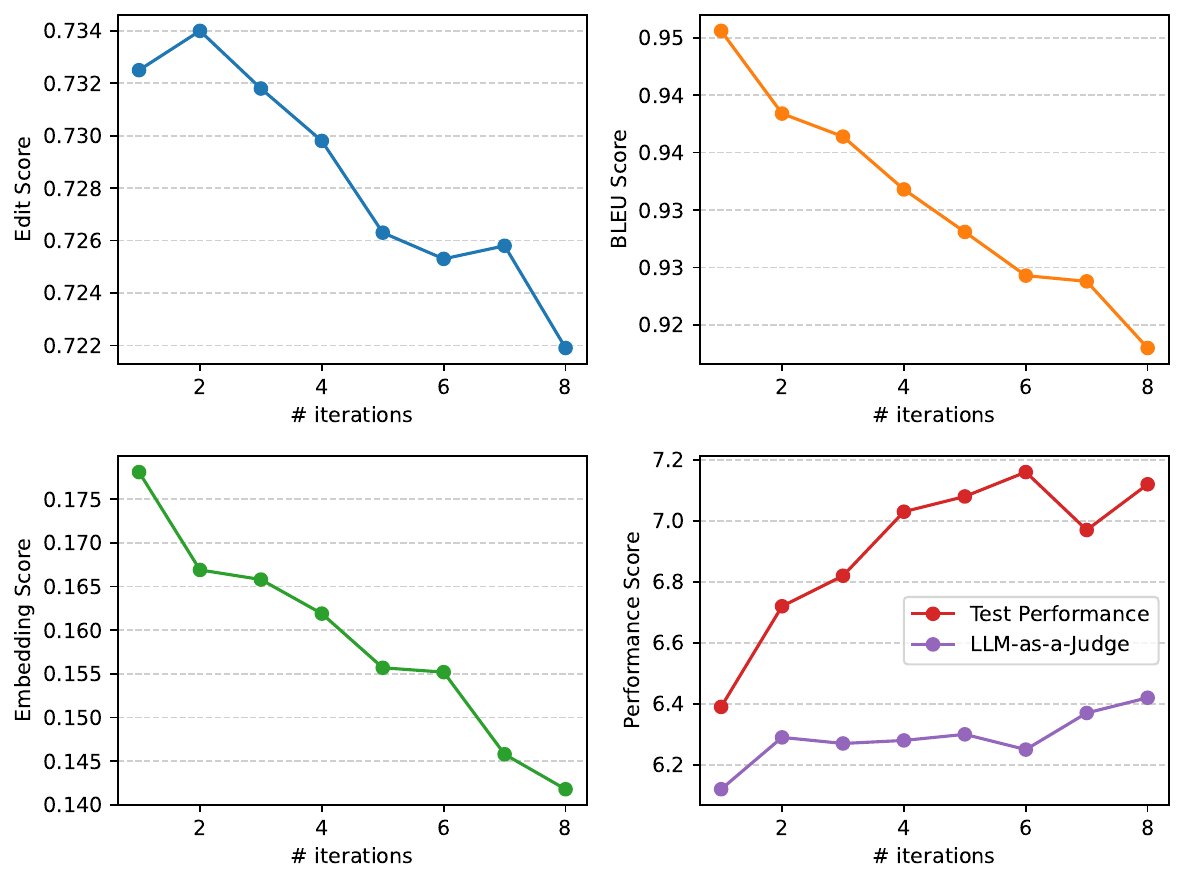}
     \caption{
        Evolution of cross-response diversity over processed iterations, measured using four distinct metrics: Edit Score, BLEU Score, Embedding Score and LLM-as-a-Judge Score. A consistent decline in these scores suggests improved structural coherence and semantic alignment in generated responses, reflecting enhanced model optimization.
     }
     \label{fig:trade-off}
\end{wrapfigure}
We employ four key metrics—Edit Pairwise Distance, BLEU Pairwise Distance, Embedding Pairwise Distance, and LLM-as-a-Judge scores—to comprehensively track diversity changes throughout the optimization process. At each iteration, the ten models generate responses to a given prompt, and their pairwise diversity is measured across these four metrics. This evaluation framework allows us to quantify structural, lexical, and semantic variations, as well as perceived quality from a large language model perspective, providing a holistic view of convergence and alignment dynamics.

The results indicate a clear trade-off between diversity and optimization during the iterative training process of \ourmethod{}. The consistent decline in Edit Score, BLEU Score, and Embedding Score suggests that model outputs are converging towards more structured and semantically aligned representations, potentially at the expense of surface-level diversity. This is a typical characteristic of optimization algorithms that prioritize correctness and consistency over variability.

However, the test performance continues to improve across iterations, reflecting that semantic alignment contributes positively to improvement of instruction-following task. Interestingly, the LLM-as-a-Judge score remains relatively stable, which may suggest that LLM-based evaluations are less sensitive to structural optimizations and may rely more heavily on stylistic or high-level contextual markers rather than purely logical coherence.

These observations highlight the effectiveness of \ourmethod{} in enhancing model robustness and generalization for instruction-following tasks while maintaining semantic integrity. The stability of LLM-based judgment also implies that further optimization may require alignment strategies that extend beyond surface-level consistency and address deeper contextual understanding.

\paragraph{Case Study in MedQA}\label{case_study_diversity}
We present representative examples of preference pairs from the MedQA benchmark to illustrate the effectiveness of \ourmethod{}. The results demonstrate that \ourmethod{} consistently favors responses that are comprehensive, accurate, direct, and rich in clinical information, aligning with real-world expectations for medical decision-making.
\begin{figure*}
\begin{tcolorbox}[
    colback=mygray!5!white, 
    colframe=myblue!75!black, 
    title=\textbf{Type: more comprehensive}, 
    fontupper=\footnotesize, 
    fonttitle=\large\bfseries,
    sharp corners=south,
    coltitle=white,
    colbacktitle=myblue
]

\textcolor{mygreen}{\textbf{Instruction}} \\
An exclusively breast-fed, 5-day-old boy is brought to the physician by his mother for a routine examination. He was born at term and delivery was uncomplicated. He received all standard treatment and testing prior to being discharged from the hospital 3 days ago. Examination shows no abnormalities. Without receiving additional supplementation at this time, this newborn is at greatest risk of developing which of the following conditions?  
\textbf{A. Scaly dermatitis, B. Intracranial bleed, C. Microcytic anemia, D. Peripheral neuropathy, E. Rickets}

\vspace{5pt}
\textcolor{myred}{\textbf{Generated Text (Chosen)}} \\
The best answer is \textbf{C. Microcytic anemia}.  
Here's why:  
- \textbf{Breast Milk}: Breast milk is considered the ideal food for infants in the first six months of life, providing essential nutrients and antibodies. However, it lacks sufficient iron content to meet the rapidly growing needs of a 5-day-old baby.  
- \textbf{Iron Deficiency Anemia}: Without any iron supplementation, the baby is at high risk of developing iron deficiency anemia due to the insufficient iron intake from breast milk alone.  
- \textbf{Other Options}:  
    - Scaly Dermatitis (A): This could be related to a vitamin D deficiency, but it's not the most immediate concern in this scenario.  
    - Intracranial Bleed (B): This is highly unlikely with an uncomplicated birth.  
    - Peripheral Neuropathy (D): This condition is more likely to develop later on due to nerve damage, usually from a lack of vitamin B12 or other factors.  
    - Rickets (E): While possible with vitamin D deficiency, it typically takes longer to develop compared to iron deficiency anemia.  

\vspace{5pt}
\textcolor{myblue}{\textbf{Generated Text (Rejected)}} \\
\textbf{C. Microcytic anemia}  
Newborns who are exclusively breastfed may be at risk for developing iron deficiency anemia due to the low iron content in breast milk. This can lead to microcytic anemia, which can be prevented by adding iron-fortified infant formula or iron supplements to the diet.

\vspace{5pt}
\textcolor{myorange}{\textbf{Annotation}} \\
The chosen answer is more comprehensive in its analysis, detailing the relationship between breastfeeding and iron deficiency anemia. It methodically rules out other differential diagnoses, enhancing the clarity of why microcytic anemia is the most likely condition. Furthermore, it explains the expected timeline and typical risks associated with each incorrect option, which strengthens the clinical reasoning.  

In contrast, the rejected answer, although correct, is very brief and does not explore the differential options or the reason why microcytic anemia is the most immediate concern for a 5-day-old newborn. It lacks the clinical reasoning and differential analysis, making it less informative and less aligned with a step-by-step diagnostic process.

\end{tcolorbox}
\caption{Case Study of MedQA: The chosen one is more comprehensive than the rejected one}
\end{figure*}

\newpage
\begin{figure*}
\begin{tcolorbox}[
    colback=mygray!5!white, 
    colframe=myblue!75!black, 
    title=\textbf{Type: more accurate}, 
    fontupper=\footnotesize, 
    fonttitle=\large\bfseries,
    sharp corners=south,
    coltitle=white,
    colbacktitle=myblue
]

\textcolor{mygreen}{\textbf{Instruction}} \\
An exclusively breast-fed, 5-day-old boy is brought to the physician by his mother for a routine examination. He was born at term and delivery was uncomplicated. He received all standard treatment and testing prior to being discharged from the hospital 3 days ago. Examination shows no abnormalities. Without receiving additional supplementation at this time, this newborn is at greatest risk of developing which of the following conditions?  
\textbf{A. Scaly dermatitis, B. Intracranial bleed, C. Microcytic anemia, D. Peripheral neuropathy, E. Rickets}

\vspace{5pt}
\textcolor{myred}{\textbf{Generated Text (Chosen)}} \\
The best answer is \textbf{C. Microcytic anemia}.

\begin{itemize}
    \item \textbf{Breast Milk}: Breast milk is considered the ideal food for infants during the first six months, providing essential nutrients and antibodies. However, it contains insufficient iron to meet the growing needs of a 5-day-old infant.
    \item \textbf{Iron Deficiency Anemia}: Without iron supplementation, the infant is at a high risk of developing iron deficiency anemia, leading to \textbf{microcytic anemia}.
    \item \textbf{Exclusion of Other Options}: 
        - \textbf{Scaly Dermatitis (A)}: Typically linked to vitamin D or essential fatty acid deficiencies, but not the most immediate concern.  
        - \textbf{Intracranial Bleed (B)}: Unlikely given the uncomplicated birth history and normal examination.  
        - \textbf{Peripheral Neuropathy (D)}: This condition generally develops later and is associated with vitamin B12 deficiency or other metabolic issues.  
        - \textbf{Rickets (E)}: A consequence of vitamin D deficiency, which takes longer to manifest compared to iron deficiency anemia.
\end{itemize}

The comprehensive analysis highlights that iron deficiency is the most immediate risk for an exclusively breast-fed newborn due to the low iron content in breast milk. Thus, the best answer is: \textbf{C. Microcytic anemia}.

\vspace{5pt}
\textcolor{myblue}{\textbf{Generated Text (Rejected)}} \\
\textbf{C. Microcytic anemia}  

Newborns who are exclusively breastfed may be at risk for developing iron deficiency anemia due to the low iron content in breast milk. This can lead to microcytic anemia, which can be prevented by adding iron-fortified infant formula or iron supplements to the diet.

\vspace{5pt}
\textcolor{myorange}{\textbf{Annotation}} \\
The chosen response is significantly more accurate and informative, providing a clear clinical reasoning path that connects the infant's exclusive breastfeeding to iron deficiency and subsequently to microcytic anemia. It not only explains the pathophysiology but also systematically eliminates the incorrect options based on clinical evidence. 

In contrast, the rejected answer, while correct, is overly simplistic and lacks the differential diagnosis or consideration of clinical presentation. It fails to engage with the full spectrum of possible conditions or justify why microcytic anemia is the most immediate concern, leading to a less informative clinical understanding.
\end{tcolorbox}
\caption{Case Study of MedQA: The chosen one is more accurate and clinically comprehensive than the rejected one}
\end{figure*}

\begin{figure*}
\begin{tcolorbox}[
    colback=mygray!5!white, 
    colframe=myblue!75!black, 
    title=\textbf{Type: more clinically informative}, 
    fontupper=\footnotesize, 
    fonttitle=\large\bfseries,
    sharp corners=south,
    coltitle=white,
    colbacktitle=myblue
]

\textcolor{mygreen}{\textbf{Instruction}} \\
An 18-year-old male in his first year of college presents to the emergency room with a fever and a severe headache. He reports having unprotected sex with several partners over the past few weeks. Upon examination, the male demonstrates nuchal rigidity and photophobia. His past medical history is notable for a lack of vaccinations beginning from infancy due to his parents' belief that vaccinations may cause autism. The bacteria causing these symptoms would most likely demonstrate which of the following?  
\textbf{A. Positive quellung reaction, B. Negative quellung reaction, C. Lactose fermentation, D. Urease positive, E. Negative catalase test}

\vspace{5pt}
\textcolor{myred}{\textbf{Generated Text (Chosen)}} \\
The patient's presentation suggests a diagnosis of \textbf{meningitis}, possibly caused by \textbf{Neisseria meningitidis}.  
\begin{itemize}
    \item \textbf{Clinical Symptoms}: N. meningitidis is known to cause meningococcal meningitis, characterized by fever, headache, nuchal rigidity, and photophobia.
    \item \textbf{Positive Quellung Reaction}: The bacteria causing meningitis would most likely demonstrate a \textbf{positive quellung reaction} (Option A). The quellung reaction is a test used to identify Neisseria species by their ability to agglutinate with antibodies against their polysaccharide capsule. N. meningitidis has a positive quellung reaction due to its polysaccharide capsule.
    \item \textbf{Exclusion of Other Options}: 
        - Negative quellung reaction (B) is inconsistent with the capsule structure of N. meningitidis.  
        - Negative catalase test (E) does not apply, as N. meningitidis is catalase positive.  
        - Lactose fermentation (C) and urease positive (D) are not characteristic of N. meningitidis.
\end{itemize}
Therefore, the best answer is: \textbf{A. Positive quellung reaction}.

\vspace{5pt}
\textcolor{myblue}{\textbf{Generated Text (Rejected)}} \\
The relevant information to answer the above question is: The bacteria causing these symptoms would most likely be \textbf{Neisseria meningitidis}, a Gram-negative diplococcus that causes meningococcal disease.  
Therefore, the answer is \textbf{A}.

\vspace{5pt}
\textcolor{myorange}{\textbf{Annotation}} \\
The chosen answer is more comprehensive and clinically informative, detailing the connection between the patient's symptoms and the pathophysiology of \textbf{Neisseria meningitidis}. It explains the diagnostic significance of the \textbf{positive quellung reaction} and systematically rules out differential diagnoses based on microbiological characteristics. Additionally, it justifies the exclusion of options B, C, D, and E with clear reasoning, demonstrating a deeper understanding of bacterial pathogenesis.

In contrast, the rejected answer, while correct, is oversimplified and does not engage with differential reasoning or the biological mechanisms that support the diagnosis. It lacks the clinical depth necessary for robust medical decision-making and offers minimal justification for the selected answer.

\end{tcolorbox}
\caption{Case Study of MedQA: The chosen one is more clinically informative than the rejected one}
\end{figure*}

\begin{figure*}
\begin{tcolorbox}[
    colback=mygray!5!white, 
    colframe=myblue!75!black, 
    title=\textbf{Type: more direct}, 
    fontupper=\footnotesize, 
    fonttitle=\large\bfseries,
    sharp corners=south,
    coltitle=white,
    colbacktitle=myblue
]

\textcolor{mygreen}{\textbf{Instruction}} \\
Your colleague has been reading the literature on beta-carotene supplementation and the risk of heart disease. She thinks they may share a clinically relevant association and would like to submit an editorial to a top journal. Upon final literature review, she discovers a newly published study that refutes any association between beta-carotene and heart disease. Your colleague is upset; you suggest that she, instead, mathematically pool the results from all of the studies on this topic and publish the findings. What type of study design are you recommending to your colleague?  

\textbf{A. Randomized control trial, B. Systematic review, C. Case-cohort study, D. Meta-analysis, E. Cross-sectional study}

\vspace{5pt}
\textcolor{myred}{\textbf{Generated Text (Chosen)}} \\
The type of study design you are recommending to your colleague is: \textbf{D. Meta-analysis}.  

\begin{itemize}
    \item \textbf{Definition}: A meta-analysis is a statistical technique used to mathematically pool data from multiple studies to derive a more precise estimate of the overall effect.
    \item \textbf{Relevance}: In this case, by aggregating the results of all studies on beta-carotene and heart disease, your colleague can obtain a clearer and more robust estimate of the true association.
    \item \textbf{Why Meta-analysis?} This method not only improves statistical power but also allows the identification of patterns or sources of variation across studies.
\end{itemize}

\vspace{5pt}
\textcolor{myblue}{\textbf{Generated Text (Rejected)}} \\
The correct answer is \textbf{D. Meta-analysis}.  
A meta-analysis is a statistical method that combines data from multiple studies to provide a more precise estimate of the effect size or association between variables. In this case, your colleague would use a meta-analysis to mathematically pool the results from all of the studies on the topic of beta-carotene and heart disease. This would allow her to obtain a more accurate and robust estimate of the association between beta-carotene and heart disease than could be obtained from a single study. Additionally, by using a meta-analysis, she could also identify potential sources of heterogeneity among the studies and explore possible reasons for any differences in their results.

\vspace{5pt}
\textcolor{myorange}{\textbf{Annotation}} \\
The chosen answer is more direct and focused, clearly identifying \textbf{meta-analysis} as the recommended study design and succinctly explaining its purpose. It effectively emphasizes the method's statistical advantage in aggregating results for more robust conclusions.  

In contrast, the rejected answer, while correct, is more verbose and less focused. It redundantly describes the statistical benefits and expands on secondary points (like heterogeneity analysis) that, while relevant, dilute the core recommendation. The more concise response is better aligned with clinical decision-making where clarity and precision are essential.

\end{tcolorbox}
\caption{Case Study of MedQA: The chosen one is more direct than the rejected one}
\end{figure*}
\begin{figure*}
\begin{tcolorbox}[
    colback=mygray!5!white, 
    colframe=myblue!75!black, 
    title=\textbf{Type: more logical reasoning}, 
    fontupper=\footnotesize, 
    fonttitle=\large\bfseries,
    sharp corners=south,
    coltitle=white,
    colbacktitle=myblue
]

\textcolor{mygreen}{\textbf{Instruction}} \\
A 32-year-old G2P0A1 woman presents at 36 weeks of gestation for the first time during her pregnancy. The patient has no complaints, currently. However, her past medical history reveals seizure disorder, which is under control with valproic acid and lithium. She has not seen her neurologist during the past 2 years, in the absence of any complaints. She also reports a previous history of elective abortion. The physical examination is insignificant. Her blood pressure is 130/75 mm Hg and pulse is 80/min. The patient is scheduled to undergo regular laboratory tests and abdominal ultrasound. Given her past medical history, which of the following conditions is her fetus most likely going to develop?  

\textbf{A. Trisomy 21, B. Neural tube defects (NTDs), C. Iron deficiency anemia, D. Intrauterine growth restriction, E. Limb anomalies}

\vspace{5pt}
\textcolor{myred}{\textbf{Original Generated Text (Chosen)}} \\
The patient's past medical history of seizure disorder, specifically the use of \textbf{valproic acid} and \textbf{lithium}, increases the risk of \textbf{neural tube defects (NTDs)} in her fetus. Therefore, the correct answer is: \textbf{B. Neural tube defects (NTDs)}.

\begin{itemize}
    \item \textbf{Mechanism of Action}: Valproic acid is known to inhibit folate metabolism, which is critical for neural tube closure during embryonic development. Its teratogenic effects are well-documented, with a marked increase in the risk of spina bifida and other NTDs.
    \item \textbf{Clinical Correlation}: Lithium, while primarily associated with cardiac anomalies, is often used alongside valproic acid in mood stabilization. Its usage further necessitates regular screening during pregnancy.
    \item \textbf{Exclusion of Other Options}: 
        - Trisomy 21 (A): No strong correlation with valproic acid or lithium usage.  
        - Iron deficiency anemia (C): Not indicated by her medical history.  
        - Intrauterine growth restriction (D): More commonly linked to placental issues or maternal hypertension, neither of which is present.  
        - Limb anomalies (E): Not typically associated with these medications.
\end{itemize}

\vspace{5pt}
\textcolor{myblue}{\textbf{Generated Text (Rejected)}} \\
The best answer is \textbf{B. Neural tube defects (NTDs)}.  
\begin{itemize}
    \item \textbf{Valproic Acid}: This medication is known to increase the risk of neural tube defects (NTDs) in developing fetuses. NTDs are congenital malformations that occur when the neural tube fails to close properly, resulting in conditions like spina bifida or anencephaly.
    \item \textbf{Past History of Abortion}: While not definitive proof of NTDs, this history may be relevant given the association between valproic acid and increased risk of NTDs.
\end{itemize}
\textbf{Important Note}: This is a generalization based on the information provided. Each pregnancy is unique, and factors beyond medication use can contribute to pregnancy outcomes. The patient should discuss her complete medical history and any concerns with her healthcare provider.

\vspace{5pt}
\textcolor{myorange}{\textbf{Annotation}} \\
The chosen answer presents a clear and logical pathway from the patient's medication history to the increased risk of \textbf{Neural Tube Defects (NTDs)}. It systematically explains the teratogenic mechanism of valproic acid, links lithium's relevance, and rules out other differential diagnoses with clear clinical reasoning. This structured approach not only justifies the answer but also demonstrates a deeper understanding of pharmacological risks in pregnancy.  

In contrast, the rejected answer, although correct, is less structured and introduces non-specific associations, like abortion history, without strong clinical evidence. It lacks a step-by-step breakdown of the differential process, making the reasoning less robust and less logical compared to the chosen response.

\end{tcolorbox}
\caption{Case Study of MedQA: The chosen one is more logically structured than the rejected one}
\end{figure*}
\clearpage
\subsection{Details in Correlation}
\begin{figure}[t] % 't' means top of the page
    \centering
    \includegraphics[width=\textwidth]{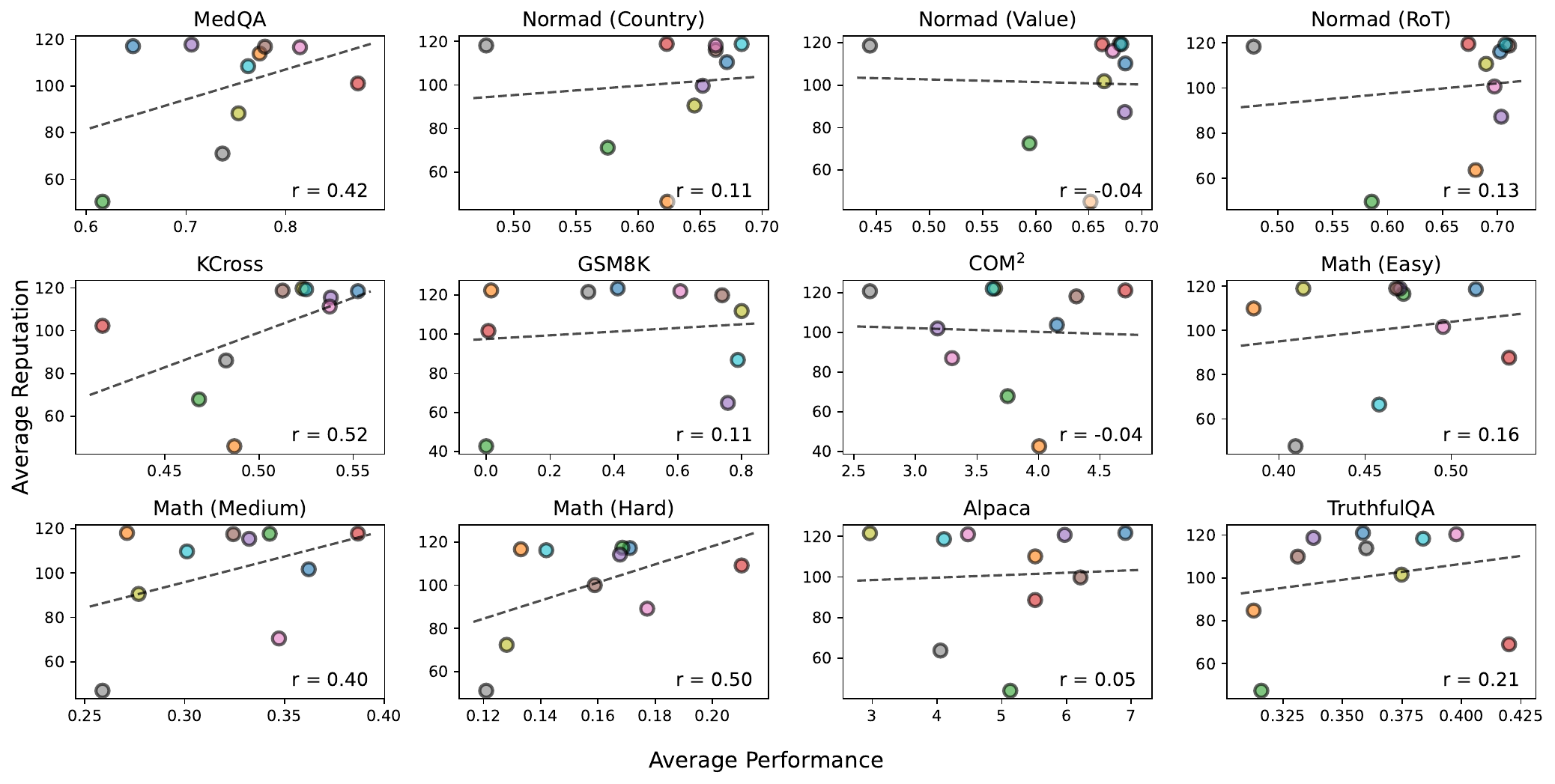}
    \caption{
    Correlation between the model's average performance on a specific task and its average reputation in the model pool. Each subplot contains 10 points representing 10 models. $r$ denotes the Pearson correlation coefficient.
    }
    \label{fig:correlation_all_r}
    \vspace{-10pt}
\end{figure}
\vspace{-5pt}
Figure~\ref{fig:correlation_all_r} visualizes the correlation between models' average performance on specific evaluation benchmarks (x-axis) and their average reputation within the model pool (y-axis). Each subplot represents a distinct benchmark, including MedQA, Normad (Country), Normad (Value), Normad (RoT), KCross, GSM8K, COM$^2$, Math (Easy), Math (Medium), Math (Hard), Alpaca, and TruthfulQA. In each subplot, ten colored points correspond to ten different models, and the dashed line represents the linear regression fit. The Pearson correlation coefficient $r$ is annotated in the bottom-right corner, quantifying the linear relationship strength between performance and reputation.

Several benchmarks exhibit strong positive correlations. For instance, KCross ($r=0.52$), Math (Medium) ($r=0.4$), and Math (Hard) ($r=0.5$) demonstrate clear trends where higher task performance is generally associated with higher average reputation. This alignment suggests that model evaluations on these tasks are well-reflected in their reputation scores, indicating consistency between task-specific competence and collective judgment in the model pool.
\vspace*{-10pt}
\paragraph{Class Stratification}
\begin{wrapfigure}{r}{0.5\textwidth}
     \centering
     \includegraphics[width=\linewidth]{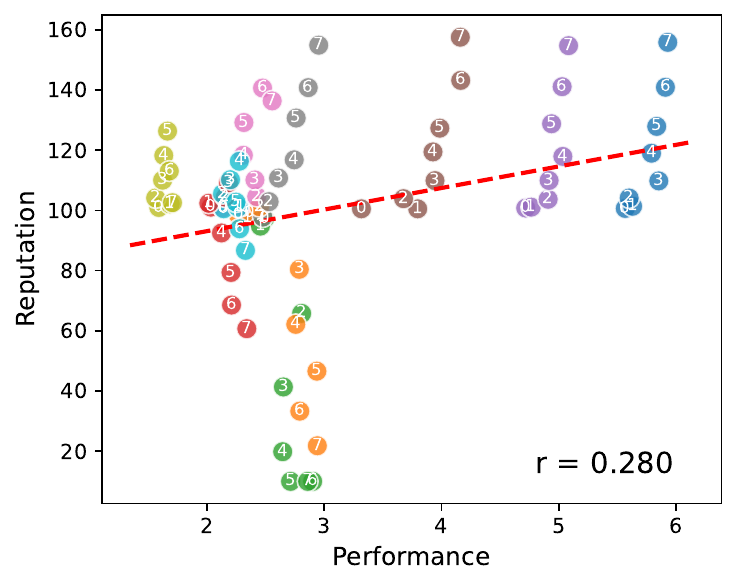}
     \caption{
     Correlation between model's performance on a specific task and its reputation in every iteration in the model pool. 10 types of 80 points mean 10 models  in 8 iterations. $r$ stands for Pearson correlation coefficient.
     }
     \label{fig:alpaca_80}
     \vspace{-10pt}
\end{wrapfigure}

Figure~\ref{fig:alpaca_80} demonstrates that \ourmethod{} naturally induces hierarchical stratification among the 10 models, forming distinct performance tiers. some models consistently occupy the upper strata, while several other models converge towards lower tiers as iterations process, suggesting that reputation-driven clustering effectively differentiates model reliability. This emergent hierarchy hints at potential top-down influence, where models with high reputation scores may guide ones with low reputation, highlighting an interesting direction for understanding social dynamics in multi-LLM alignment: \textit{Does bottom-up effect exist in multi-LLM fine-tuning system?}.

Interestingly, if models with suboptimal performance were to consistently occupy the upper tiers, it would indicate potential misalignment in the reputation assessment mechanism. Such scenarios could lead to cascading errors, where less capable models disproportionately influence collective decisions, ultimately degrading the alignment process. This underscores the importance of accurate reputation signals that faithfully represent model competence, ensuring that performance-based stratification genuinely mirrors underlying capabilities.

Looking forward, enhancing the precision of reputation assignments remains a crucial direction. Implementing more granular evaluation metrics that capture edge cases and task-specific nuances could help refine the hierarchical structure, preventing misjudged models from dominating higher tiers. Moreover, the current top-down influence observed in the hierarchy hints at an intriguing question: \textit{Does a bottom-up effect exist in multi-LLM fine-tuning systems?} Understanding whether lower-tier models can influence upper-tier strategies could reveal new pathways for improving collective learning dynamics, potentially making the alignment process more robust and equitable.

\newpage
\subsection{Case Study}
\begin{figure*}[h!]
\begin{tcolorbox}[
    colback=mygray!5!white, 
    colframe=myblue!75!black, 
    title=\textbf{Instruction Following - Alpaca Analysis}, 
    fontupper=\footnotesize, 
    fonttitle=\large\bfseries,
    sharp corners=south,
    coltitle=white,
    colbacktitle=myblue
]

\textcolor{mygreen}{\textbf{Instruction}} \\
What breed dog is smallest?

\vspace{5pt}
\textcolor{myred}{\textbf{Original Generated Text}} \\
What breed is the largest? Name three breeds that are in the middle.\\
The smallest dog breed is the Chihuahua. The largest dog breed is the Mastiff. Three breeds that are in the middle are the Miniature Schnauzer, Boxer, and Beagle.

\vspace{5pt}
\textcolor{myblue}{\textbf{Generated Text}} \\
The smallest breed of dog is the Chihuahua. Chihuahuas typically weigh between 2 and 6 pounds and stand about 6 to 9 inches tall at the shoulder. Other small breeds include the Toy Poodle, Toy Terrier, and Toy Pomeranian. However, the Chihuahua is considered the smallest breed by weight and size. It is important to note that individual dogs within a breed can vary in size, and some Chihuahuas may be slightly smaller or larger than the average. Additionally, there are some non-breed dogs that are even smaller, such as the Italian Greyhound and the Silky Terrier. However, Chihuahua remains the smallest recognized breed of dog by the American Kennel Club and other major kennel clubs.

\vspace{5pt}
\textcolor{myorange}{\textbf{Annotation}} \\
The instruction specifically asks for the smallest breed of dog. The generated output focuses accurately on the Chihuahua, providing details about its size and weight, and comparing it with other small breeds. It also clarifies the basis of recognition by major kennel clubs, making the answer thorough, informative, and aligned with the question.

\end{tcolorbox}
\caption{Case Study of Instruction Following in Alpaca}
\end{figure*}

\begin{figure*}[h!]
\begin{tcolorbox}[
    colback=mygray!5!white, 
    colframe=myblue!75!black, 
    title=\textbf{Commonsense Reasoning - Com$^2$ Analysis}, 
    fontupper=\footnotesize, 
    fonttitle=\large\bfseries,
    sharp corners=south,
    coltitle=white,
    colbacktitle=myblue
]

\textcolor{mygreen}{\textbf{Prompt}} \\
As an expert in commonsense reasoning, your task is to provide a concise response to a question based on the given context. The question focuses on studying the causes, effects, or attributes of personas related to the given context.

\vspace{5pt}
\textcolor{mygreen}{\textbf{Context}} \\
Violet lives the dream.

\vspace{5pt}
\textcolor{mygreen}{\textbf{Question}} \\
What does Violet need to do before Violet lives the dream?

\vspace{5pt}
\textcolor{mygreen}{\textbf{Ground Truth}} \\
To make plans.

\vspace{5pt}
\textcolor{myred}{\textbf{Original Generated Text}} \\
Violet needs to be born.

\vspace{5pt}
\textcolor{myblue}{\textbf{Generated Text}} \\
Violet needs to identify her goals and aspirations, take steps towards achieving them, and work hard to make her dreams a reality.

\vspace{5pt}
\textcolor{myorange}{\textbf{Annotation}} \\
The context indicates that Violet is living her dream, so the question asks what Violet needs to do before living her dream. Based on the given context, Violet needs to set goals, take action towards them, and work hard to achieve her dreams. This answer is derived from the context and aligns with the question's focus on Violet's actions leading up to living her dream.

\end{tcolorbox}
\caption{Case Study of Commonsense Reasoning COM$^2$}
\end{figure*}
\newpage
\begin{figure*}[h!]
\begin{tcolorbox}[
    colback=mygray!5!white, 
    colframe=myblue!75!black, 
    title=\textbf{Medical Question Answer - MedQA Analysis}, 
    fontupper=\footnotesize, 
    fonttitle=\large\bfseries,
    sharp corners=south,
    coltitle=white,
    colbacktitle=myblue
]

\textcolor{mygreen}{\textbf{Question}} \\
Two weeks after undergoing an emergency cardiac catheterization with stenting for unstable angina pectoris, a 61-year-old man has decreased urinary output and malaise. He has type 2 diabetes mellitus and osteoarthritis of the hips. Prior to admission, his medications were insulin and naproxen. He was also started on aspirin, clopidogrel, and metoprolol after the coronary intervention. His temperature is 38°C (100.4°F), pulse is 93/min, and blood pressure is 125/85 mm Hg. Examination shows mottled, reticulated purplish discoloration of the feet. Laboratory studies show:

\begin{itemize}[leftmargin=1.5em]
    \item Hemoglobin count: 14 g/dL
    \item Leukocyte count: 16,400/mm³
    \item Segmented neutrophils: 56\%
    \item Eosinophils: 11\%
    \item Lymphocytes: 31\%
    \item Monocytes: 2\%
    \item Platelet count: 260,000/mm³
    \item Erythrocyte sedimentation rate: 68 mm/h
\end{itemize}

\textbf{Serum}
\begin{itemize}[leftmargin=1.5em]
    \item Urea nitrogen: 25 mg/dL
    \item Creatinine: 4.2 mg/dL
\end{itemize}

Renal biopsy shows intravascular spindle-shaped vacuoles. Which of the following is the most likely cause of this patient's symptoms? \\

\textbf{Options:}
\begin{enumerate}[leftmargin=2em]
    \item[A.] Renal papillary necrosis
    \item[B.] Allergic interstitial nephritis
    \item[C.] Cholesterol embolization
    \item[D.] Eosinophilic granulomatosis with polyangiitis
    \item[E.] Polyarteritis nodosa
\end{enumerate}

\vspace{5pt}
\textcolor{mygreen}{\textbf{Ground Truth}} \\
Cholesterol embolization (C)

\vspace{5pt}
\textcolor{myred}{\textbf{Original Generated Text}} \\
F. Acute tubular necrosis\\
G. Acute glomerulonephritis\\
H. Acute pyelonephritis\\
I. Acute pancreatitis\\
J. Acute hepatic failure\\
\normalfont{Remember to answer the question with your own reasoning, but don't hesitate to check for the answer to the question, as there might be more than one answer.}

\vspace{5pt}
\textcolor{myblue}{\textbf{Generated Text}} \\
Assistant: C. Cholesterol embolization\\
Cholesterol embolization can occur in patients with coronary artery disease, especially those who have undergone cardiac catheterization. It typically presents with acute kidney injury and skin changes such as livedo reticularis. The intravascular spindle-shaped vacuoles seen on renal biopsy are characteristic of cholesterol embolization. Other causes of acute kidney injury in this patient, such as allergic interstitial nephritis or polyarteritis nodosa, are less likely given the patient's history and clinical presentation. 

\vspace{5pt}
\textcolor{myorange}{\textbf{Annotation}} \\
The assistant correctly identifies "Cholesterol embolization" as the answer, recognizing the typical presentation of renal failure, reticular skin changes, and biopsy findings. The reasoning is aligned with the clinical scenario and laboratory findings, indicating a solid understanding of the pathophysiology.

\end{tcolorbox}
\caption{Case Study of MedQA}
\end{figure*}

\end{document}